\documentclass[sigplan,screen]{acmart}
\AtBeginDocument{%
  }

\setcopyright{cc}
\setcctype{by}
\acmDOI{10.1145/3759425.3763389}
\acmYear{2025}
\copyrightyear{2025}
\acmISBN{979-8-4007-2148-9/25/10}
\acmConference[LMPL '25]{Proceedings of the 1st ACM SIGPLAN International Workshop on Language Models and Programming Languages}{October 12--18, 2025}{Singapore, Singapore}
\acmBooktitle{Proceedings of the 1st ACM SIGPLAN International Workshop on Language Models and Programming Languages (LMPL '25), October 12--18, 2025, Singapore, Singapore}
\received{2025-06-28}
\received[accepted]{2025-08-08}

\usepackage{fvextra}
\usepackage{fancyvrb}
\usepackage{latexsym}
\usepackage{wasysym}
\usepackage{xcolor}
\usepackage{subcaption}
\usepackage[breakable]{tcolorbox}
\usepackage{enumitem}
\usepackage{xcolor}
\usepackage{listings}
\usepackage{multirow}
\usepackage{microtype} 
\usepackage{multicol}
\usepackage{balance}
\begin{document}

\title{Understanding Formal Reasoning Failures in LLMs as Abstract Interpreters}

\author{Jacqueline Mitchell}
\affiliation{%
  \institution{University of Southern California}
  \city{Los Angeles}
  \country{USA}}
\email{jlm41510@usc.edu}

\author{Brian Hyeongseok Kim}
\affiliation{%
  \institution{University of Southern California}
  \city{Los Angeles}
  \country{USA}}
\email{brian.hs.kim@usc.edu}

\author{Chenyu Zhou}
\affiliation{%
 \institution{University of Southern California}
 \city{Los Angeles}
 \country{USA}}
\email{czhou691@usc.edu}

\author{Chao Wang}
\affiliation{%
 \institution{University of Southern California}
 \city{Los Angeles}
 \country{USA}}
\email{wang626@usc.edu}

\renewcommand{\shortauthors}{Mitchell et al.}

\begin{abstract}
    Large language models (LLMs) are increasingly used for program verification, and yet little is known about \emph{how} they reason about program semantics during this process.  In this work, we focus on abstract interpretation based-reasoning for invariant generation and introduce two novel prompting strategies that aim to elicit such reasoning from LLMs.  We evaluate these strategies across several state-of-the-art LLMs on 22 programs from the SV-COMP benchmark suite widely used in software verification.  We analyze both the soundness of the generated invariants and the key thematic patterns in the models' reasoning errors.  This work aims to highlight new research opportunities at the intersection of LLMs and program verification for applying LLMs to verification tasks and advancing their reasoning capabilities in this application.
\end{abstract}



\settopmatter{printacmref=true}
\maketitle

\section{Introduction}
LLMs have undeniably changed the way we interact with software, from development to analyzing programs.  
A particularly salient use case for researchers in program analysis has been leveraging LLMs as oracles to generate invariants for program analysis, leading to a plethora of research papers on the topic \cite{chakraborty2023ranking, wu2024llm, li2024enhancing, liu2024llmse, DBLP:journals/corr/abs-2311-07948}.  
They have shown that LLMs have the potential to generate invariants when incorporated into a verification pipeline.
However, there is a gap in the literature studying \emph{the reasoning process} that LLMs take to generate these invariants. 
Despite the impressive coding ability of LLMs, it remains an open problem whether LLMs are able to formally reason about program semantics, or if the generated invariants are derived fortuitously by looking at the code's structure.  
Recent work on evaluating LLM performance on various tasks, such as control, planning, and dealing with abstracted versions of common reasoning problems (e.g., logical puzzles), suggests that their reasoning abilities fall short, particularly when not supported by external verification tools \cite{DBLP:conf/icml/KambhampatiVGVS24}.
Building upon this concern, our work shifts focus from the program invariants LLMs generate and their correctness to the reasoning process behind invariant generation, aiming to identify common errors and limitations in the models' internal logic.

To this end, we choose to explore the reasoning process through the lens of abstract interpretation \cite{DBLP:conf/popl/CousotC77}, a systematic framework capable of effectively generating program invariants by soundly over-approximating the semantics of programs.
Our motivation for taking an abstract interpretation-based perspective is that it provides a sequence of verifiable steps used to derive program invariants.
If asked to explain a program invariant it generated, an LLM will most likely provide an \emph{natural language} explanation (e.g., ``$x > 2$ because $x$ is larger than one prior to adding $1$ to it'').
Such explanations can become increasingly difficult to verify, especially as programs become more complex in their semantics.
Thus, we prompt LLMs to reason in the style of abstract interpretation during the invariant generation task so that we can check each step of the process.
Although it could be possible to retrain or fine-tune a model to improve its reasoning capabilities in this area, this may not be feasible for many users.
Thus, we opt for prompting-based evaluation to audit the abstract interpretation-based reasoning of a given model and provide some insight on its reasoning capability.

To facilitate this, we introduce two novel prompting strategies denoted as \emph{Compositional} and \emph{Transitional} to elicit the two different core styles \cite{rival2020introduction} of abstract interpretation-based reasoning.
We opt for a prompting-based strategy, as opposed to mechanistic interpretability techniques, for several reasons.
The first reason is so that the methodology is applicable to any model, closed or open source, as many of the models used for program invariant generation (e.g., ChatGPT, Claude) are closed source.
Similarly, directly observing the internal mechanisms of these models can be difficult, we instead probe their reasoning indirectly by designing prompting strategies that elicit step-by-step explanations alongside the final abstract states.

The contributions of this paper are as follows: \textbf{(1)} We introduce two prompting strategies based on the two styles of abstract interpretation to generate step-by-step reasoning traces for evaluation; \textbf{(2)} We evaluate the prompting strategies across four state-of-the-art LLMs on 22 SV-COMP benchmarks that are widely used in program verification; \textbf{(3)} We provide a thematic error analysis of the reasoning mistakes made by the LLMs and highlight opportunities for future researchers.

The rest of the paper is organized as follows. 
Section~\ref{section:background} goes over the preliminaries and related work. 
Section~\ref{section:program} formulates our research problem. 
Section~\ref{section:prompting} presents our prompting techniques for invariant generation. 
Section~\ref{section:evaluation} shows our main evaluation results. 
Section~\ref{section:error_analysis} highlights key thematic errors we observed in LLMs.
Section~\ref{section:conclusion} concludes our paper.

\section{Background and Related Work}
\label{section:background}

\subsection{Large Language Models}
Large language models (LLMs) are language models that have been trained on vast and diverse corpora spanning natural languages and programming languages \cite{DBLP:journals/corr/abs-2307-09288, DBLP:journals/corr/abs-2308-12950, DBLP:journals/corr/abs-2303-08774, DBLP:conf/nips/VaswaniSPUJGKP17}. 
These models are now widely accessible in various forms ranging from open-source implementations, publicly released weights, to closed-source systems.
The availability of such models, along with their increasing capability in variety of skills, has led to a substantive body of work leveraging LLMs for various reasoning tasks; formal verification and program analysis have been no exception.
As LLMs become increasingly capable, researchers have explored their potential for aiding formal reasoning and program analysis.

\paragraph{LLM-aided Program Analysis} 
Recent work has explored integrating LLMs into program analysis to enhance semantic understanding and automation beyond traditional static analysis techniques. 
Applications span bug detection, vulnerability analysis, and verification.
LLMDFA~\cite{wang2024llmdfa} proposes a compilation-free dataflow analysis framework that uses few-shot prompting, expert tools, and structured decomposition to avoid hallucinations. 
LLift~\cite{li2024enhancing} augments static analysis with LLMs to detect use-before-initialization bugs, improving path sensitivity with constraint-guided reasoning.
LATTE~\cite{liu2025llm} applies LLMs to static binary taint analysis, combining prompt engineering with slicing to track data flows in compiled code. 
Though these tools have shown the potential for LLMs to understand and handle dataflow relations, they do not evaluate the actual \emph{reasoning} capability of LLMs.  
Specifically, it is unclear if LLMs have a deep understanding of code semantics, or if positive results arise from pattern-matching on the syntax of the code.

\paragraph{LLM-aided Invariant Generation}

Recent research has also explored using LLMs to infer program invariants, offering a new avenue beyond traditional static invariant generation techniques.
For example, Automated Program Refinement~\cite{cai2025automated} uses LLMs guided by refinement calculus to generate candidate specifications including loop invariants, and verifies them using formal proof engines. Other approaches treat LLMs as black-box invariant generators and apply neuro-symbolic filtering pipelines. For example, Chakraborty et al.~\cite{chakraborty2023ranking} use LLMs to generate loop invariant candidates and propose iRank, a neural ranking model to prioritize verifiable candidates. Wu et al.~\cite{wu2024llm} propose a generate–filter–verify loop using LLMs for candidate generation, bounded model checking (BMC) for filtering, and theorem provers for final verification, solving 90\% of benchmarks in classic invariant synthesis suites. LLM-SE~\cite{liu2024llmse} combines symbolic execution with LLM reasoning to infer invariants over heap-manipulating programs, showing strong results on the LIG-MM benchmark. ACInv~\cite{liu2024enhancing} augments LLMs with static analysis summaries and introduces LLM-based refinement steps to iteratively strengthen or fix invalid invariants. Across these efforts, LLMs demonstrate high generalization, while formal tools provide rigor and soundness guarantees, which forms a promising hybrid strategy for invariant generation.
In these studies, LLMs were capable of generating invariants, but there was no investigation on \emph{how} the invariants were being generated by the LLMs, and if they are capable of generating the invariants in a way where the soundness of the reasoning process is checkable (e.g., examining the traces of an abstract interpreter).
That is, it is unclear if the reasoning process of the LLMs itself, is verifiable.

\paragraph{Understanding Internal Reasoning in LLMs}  
Outside of the context of program analysis, there is much work aimed at understanding the internal reasoning processes of LLMs.  For open weight models, these techniques include mechanistic interpretability techniques (e.g., circuit analysis, probing, activation patching) \cite{DBLP:conf/nips/ConmyMLHG23, DBLP:journals/tmlr/BereskaG24, DBLP:journals/corr/abs-2404-15255}, gradient-based methods (e.g., integrated gradients) \cite{DBLP:journals/corr/abs-2412-03886}, and representation analysis (e.g., steering vectors) \cite{DBLP:journals/corr/abs-2412-07334, DBLP:conf/iclr/0003ZSPYD25, DBLP:journals/corr/abs-2502-19721}.  For closed source models, the available techniques for understanding reasoning are largely prompt-driven \cite{DBLP:journals/corr/abs-2410-13073, DBLP:journals/corr/abs-2308-06834, DBLP:journals/corr/abs-2405-06703}.  Many of the LLMs achieving state-of-the-art results in program verification tasks, such as GPT-4o \cite{openai2024gpt4o} are closed source, motivating our prompt-driven approach in this work.

\subsection{Abstract Interpretation}

Abstract interpretation is a foundational static program analysis technique which proves properties about programs by approximating their semantics \cite{DBLP:conf/popl/CousotC77}.
In contrast to model-checking techniques like CEGAR \cite{DBLP:journals/jacm/ClarkeGJLV03} which focus on checking a program against a provided input specification, abstract interpretation can automatically compute program invariants.
More specifically, it can be used to compute program invariants for each location within a program, which we denote  as program Invariant Maps (IMs):
\begin{definition}[Program Invariant Map (IM)]
    An invariant map $\phi : Loc \rightarrow \mathcal{A}$ maps each program location $\ell$ to an invariant described in the logic of the chosen abstract domain, $\mathcal{A}$.
\end{definition}

An \emph{abstract domain} describes and approximates program semantics within some logic.
A commonly used, yet easy-to-understand abstract domain is the \emph{interval domain}. It is a non-relational domain where each integer program variable is overapproximated by an interval. (Note that it is also standard to view this domain as the set of maps from program variables to intervals.)

\begin{definition}[Interval Domain]
    The interval domain can be denoted as $\langle \mathcal{I}, \sqsubseteq, \sqcup, \sqcap \rangle$, 
    where the carrier set $\textstyle \mathcal{I} := \{[a, b] \\ ~|~ a, b \in \mathbb{Z} \cup \{-\infty, \infty\} \wedge a \leq b \} \cup \{\bot\}$,
    and $\bot$ corresponds to the empty interval.  
    The partial order $(\sqsubseteq)$, join $(\sqcup)$, and meet $(\sqcap)$, are standardly defined \cite{DBLP:conf/popl/CousotC77}.
\end{definition}

While many other domains exist for numerical invariants, ranging from convex polyhedra \cite{DBLP:conf/popl/CousotH78} to pentagons \cite{DBLP:journals/scp/LogozzoF10},
we focus on the interval domain in this work due to the simplicity of its operations.  This allows us to concentrate on how LLMs perform abstract interpretation at a conceptual level, rather than getting entangled in the complexity of specific domain operations, such as convex hull computations required by the convex polyhedra domain.
This choice is further motivated by the potential for LLMs to eventually offload such complex operations to external libraries.

For a program that manipulates integer program variables, each operation in the concrete domain has a corresponding abstract operation in the abstract domain, called an \emph{abstract transformer}.  
For instance, integer addition ($+$) in the interval domain ($+^\sharp$) corresponds to $[a,b] +^\sharp [c, d] = [a+c, b+d]$. 
For a more exhaustive description of abstract transformers, we refer the reader to standard sources \cite{DBLP:conf/popl/CousotC77, DBLP:conf/popl/CousotH78}.

Control flows necessitate \emph{join} ($\sqcup$) operations to ensure that we account for all control flow paths soundly.
For example, if the abstract state at the end of the body of an if-branch is $[0,3]$ and at the end of the body of an else-branch is $[2,4]$,
then, the two intervals must be joined to represent the abstract state at the end of the entire if-then-else block. That is, $[0,3] \sqcup [2,4] = [0,4]$.

Conditional guards necessitate \emph{meet} ($\sqcap$) operations to filter and restrict the abstract state.
For example, if $x$ is represented by $[0,6]$ in the interval domain, entering the conditional guard $x \leq 4$ will restrict it to $[0,4]$. That is, $[0,6] \sqcap [-\inf, 4] = [0,4]$.

Loops in programs necessitate fixpoint computation, often with the help of \emph{widening} ($\nabla$) operations. For example, a potentially infinitely ascending chain (resulting from an unbounded number of loop iterations) necessitates the use of a widening operator to enforce termination.
Given two intervals  where $[a,b]$ represents the value of a variable in one loop iteration and $[c,d]$ represents the value in the next loop iteration,  a possible widening operator to prevent the value from diverging endlessly is: $[a, b] \nabla [c, d] = [\text{ite}(c < a, -\infty, a), \text{ite}(b < d, \infty, d)]$.  Here, the function $\text{ite}(i_1, i_2, i_3)$ stands for If-Then-Else, meaning ``if ($i_1$) then $i_2$ else $i_3$''.

In sum, key elements required for computing invariant maps using abstract interpretation include: 
(1) abstract transformers for a given abstract domain, 
(2) join operations to soundly combine control flows, 
(3) 
meet operations used to restrict the abstract state to account for a conditional guard, 
and 
(4) widening operators to perform fixpoint computation.

\noindent \paragraph{Two Flavors of Abstract Interpretation} Using the aforementioned key elements, abstract interpreters have two distinct flavors: the \emph{compositional} perspective and the \emph{transitional} perspective \cite{rival2020introduction}.  In the former, each program construct is interpreted as a mathematical object, where the abstract semantics are defined inductively on the syntax of the program structures (e.g., while loops, if-then-else statements, and sequential composition).  In contrast, the latter perspective interprets the program as a control flow graph and uses techniques such as chaotic iterations on a system of equations representing the program semantics until convergence \cite{DBLP:conf/ershov/Bourdoncle93}. These two distinct views on abstract interpretation inspire our design of the two prompting strategies introduced in this paper. 
In the section that follows, we describe the format of the programs we consider, for use as inputs to the two prompting strategies.

\begin{figure}
    \footnotesize
    \centering
    {
    \begin{align*}
        E &:= n \mid x \mid E \astrosun E \mid \mathtt{read()} \\
        B &:= x \oplus n \text{ }  \mid \text{ }  B \&\& B \text{ }  \mid \text{ } B \mid\mid B \text{ }  \mid \text{ } !B \mid \text{ } (B) \\
        C &:= \mathtt{skip} \mid C ; C \mid x := E \mid \\ & \mathtt{if}(B) \text{ then } \{C\} \text{ else }\{C\}  \text{ end } \mid \mathtt{while} \text{ do } (B) \{C\} \text{ end } \\
        P &:= C \\
        \astrosun &\in \{+, -, *, /\} \text{  (arithmetic operators)  } \\  
        \oplus &\in \{<, <=, ==, >, >=\} \text{ (comparison operators) } \\ 
        n &\in \mathbb{Z} \text{  (integers)  } \\
        x &\in \mathbb{X} \text{ (program variables) }
    \end{align*}
     }%
    \caption{A simple IMP-like context-free grammar for some program $P$.}
    \label{fig:imp-grammar}
\end{figure}

\lstset{
    escapeinside={/*@}{@*/},
    moredelim=[is][\color{red}]{@R@}{@RE@},     
    moredelim=[is][\color{blue}]{@B@}{@BE@},    
    moredelim=[is][\color{green}]{@G@}{@GE@}    
}

\begin{figure}[t]
    \centering
    \lstset{
        basicstyle=\scriptsize\ttfamily,
        basewidth={0.5em,0.5em},   
        lineskip=0pt              
    }
    \begin{tcolorbox}[width=\columnwidth, left=1pt, right=1pt, top=1pt, bottom=1pt, boxsep=0pt, arc=0pt]
        \begin{minipage}{0.48\linewidth} 
            \begin{lstlisting}
        @B@{P0}@BE@
        /*@$a := read();$@*/
        @B@{P1}@BE@
        if (a > 6) then
          @R@[if_then]@RE@
          @B@{P2}@BE@
          /*@$a := 0;$@*/
          @B@{P3}@BE@
        else
          @R@[if_else]@RE@
          @B@{P4}@BE@
          skip;
          @B@{P5}@BE@
        end @R@[endif]@RE@
            \end{lstlisting}
        \end{minipage}
        \hfill
        \begin{minipage}{0.48\linewidth} 
            \begin{lstlisting}
@B@{P6}@BE@
while (a < 6) do
  @R@[while_true]@RE@
  @B@{P7}@BE@
  /*@$a := a + 1$;@*/
  @B@{P8}@BE@
end @R@[while_false]@RE@
@B@{P9}@BE@
            \end{lstlisting}
        \end{minipage}
    \end{tcolorbox}
    \caption{Annotated program written in the grammar of Figure~\ref{fig:imp-grammar}. \textcolor{blue}{\{P0\}} ... \textcolor{blue}{\{P9\}} mark program locations and \textcolor{red}{[directives]} mark control flow, which are included to help LLMs identify where to compute abstract states.}
    \label{fig:annotated_program}
\end{figure}

\begin{figure*}[t]
    \centering
    \includegraphics[width=\linewidth]{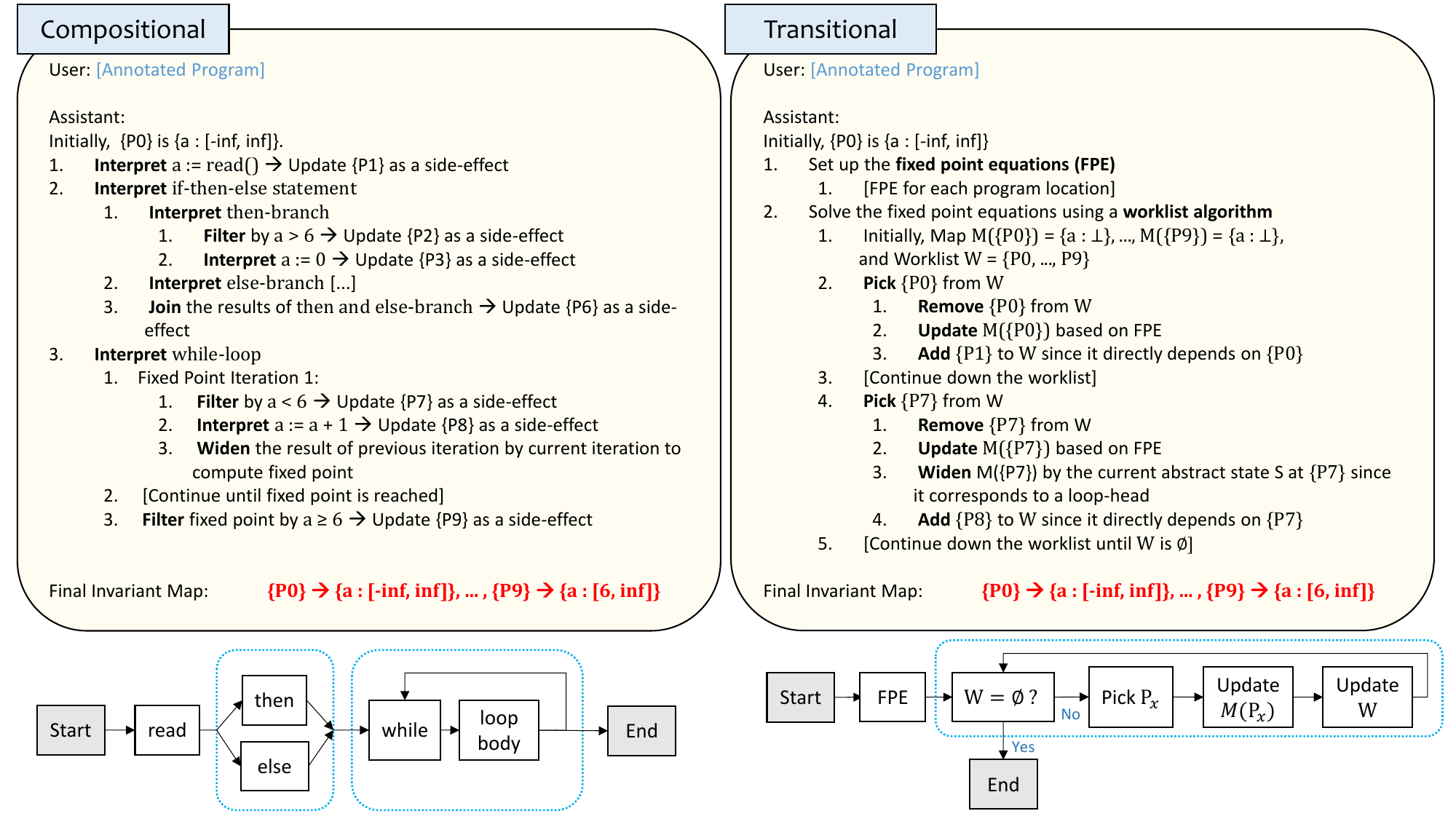}
    \caption{Two strategies for abstract interpretation: Compositional (left) and Transitional (right). 
    The annotated program from Figure~\ref{fig:annotated_program} is given as the in-context input, and the texts above show the in-context outputs corresponding to the two strategies. The flowcharts on the bottom visually represent the algorithmic flow of the two strategies.}
    \label{fig:prompts}
\end{figure*}

\section{Program Format}
\label{section:program}

This section describes how we represent and annotate programs to prepare them as input for LLMs in the context of our work.  We focus on a minimal language and control-flow annotations to isolate reasoning behavior from any learned familiarity with real-world programming languages.

\subsection{Program Representation}

In this work, we consider integer-valued programs, expressed in a simple intermediate representation (IR) language similar to IMP, expressed in the context free grammar featured in Figure ~\ref{fig:imp-grammar}.  We use a simple IR not only because it simplifies the program representation, but also because language models have been largely trained on code generated by real programming languages such as C, Python, or TypeScript \cite{DBLP:conf/emnlp/ShenCWSS22, DBLP:conf/icse/IzadiKDOPD24}.  Our goal is to eliminate any potential advantage LLMs may gain from analyzing programs written in a familiar language, as recent research suggests that their performance on various tasks can be artificially inflated by the familiarity of the problem’s representation \cite{DBLP:conf/icml/KambhampatiVGVS24}.

\subsection{Program Annotations}

In our work, we assume that programs are annotated to help LLMs better understand the structure of the program.  An example of an annotated program is shown in Figure~\ref{fig:annotated_program}.  The \textcolor{blue}{blue} annotations represent \emph{program locations}, and the \textcolor{red}{red} annotations represent \emph{control flow directives}.
This is done with the goal of preventing LLMs from making trivial mistakes in tasks such as labeling program points or identifying program structures (e.g., while loops, if-then-else statements).

\textbf{Program locations.} Program location \textcolor{blue}{$\{P0\}$} marks the beginning of the program. In the case of assignment or skip statements, a program location appears after each one.  
In the case of if-then-else ($\mathtt{if-}$) statements, a program location appears at the beginning of the then and else branches, as well as after the entire $\mathtt{if}$-statement.  
For $\mathtt{while}$-loops, a program location appears just before the body of the loop, and after the loop itself.

\textbf{Control flow directives} indicate explicit control flow structure of the program.  To reason soundly about program behavior at control-flow-sensitive program locations, such as $\mathtt{if}$-statements and $\mathtt{while}$-loops, abstract interpretation relies on operations like join, filtering, and widening.  While a human can infer control flow from syntax alone, LLMs may struggle with such structural understanding.  Thus, we annotate the program with directives that make these relationships transparent to the model.

\textbf{$\mathtt{if}$-statements} rely on the join $(\sqcup)$ operation at the end of the statements to soundly over-approximate all possible program behaviors.
Using the example from Figure~\ref{fig:annotated_program}, 
the abstract state at \textcolor{blue}{\{P6\}} is the result of joining the abstract states at \textcolor{blue}{\{P3\}} and at \textcolor{blue}{\{P5\}}.
This is indicated to the LLM with an \textcolor{red}{$[\mathtt{endif}]$} directive.
The branching statements depend on a filtering operation to satisfy the condition to enter either branch.  
Looking at Figure~\ref{fig:annotated_program} again, 
suppose that the abstract state at \textcolor{blue}{\{P1\}} is $a \mapsto [-10, 20]$.
Then, the abstract state at \textcolor{blue}{\{P2\}} is $a \mapsto [7, 20]$, as the condition indicates that $a > 6$.
This is indicated to the LLM with a \textcolor{red}{$[\mathtt{if\_then}]$} directive.  Analogously, the same thing is done in the else branch with the negation of the condition.  This is indicated with an \textcolor{red}{$[\mathtt{if\_else}]$} directive.

\textbf{$\mathtt{while}$-loops} also rely on the join and filtering operations.  
The abstract state immediately before the while loop  
(\textcolor{blue}{\{P6\}})
and the abstract state immediately after the last statement of the loop body (\textcolor{blue}{\{P8\}})
are joined together for both the program locations at the loop head (\textcolor{blue}{\{P7\}}) and after the loop (\textcolor{blue}{\{P9\}}). 
To account for the loop guard being true (\textcolor{blue}{\{P7\}}), the join operation is filtered by the loop guard; the LLM is instructed to do this with a \textcolor{red}{$[\mathtt{while}\_\mathtt{true}]$} directive.  
To account for the false case (\textcolor{blue}{\{P9\}}), the join operation is filtered by the negation of the loop guard; this is indicated to the LLM with a \textcolor{red}{$[\mathtt{while}\_\mathtt{false}]$} directive. The
\textcolor{red}{$[\mathtt{while}\_\mathtt{true}]$} directive also indicates to the LLM to perform widening to accelerate the convergence of fixpoint computation.

\section{Prompting Techniques}
\label{section:prompting}

With the program encoding and annotations defined, we now introduce our two prompting strategies for abstract interpretation. These correspond to two distinct approaches: the \textbf{Compositional} Strategy and the \textbf{Transitional} Strategy.

Given that both flavors of abstract interpretation are inherently algorithmic, we design our prompts in a manner inspired by the Algorithm of Thoughts (AoT)~\citep{DBLP:journals/corr/abs-2308-10379} technique. 
AoT is similar to Chain of Thought~\citep{DBLP:conf/nips/Wei0SBIXCLZ22} but further integrates the search process into their few-shot learning~\cite{DBLP:journals/corr/abs-2005-14165}.
In-context examples in AoT are designed to illustrate how to evaluate each solving step. This is meant to guide the LLM and help it decide whether it should explore a problem subtree further or backtrack to find a different viable subtree to make progress towards the solution.  
Our full prompts for the two strategies are provided the Appendix of the extended version\footnote{https://arxiv.org/abs/2503.12686} of our paper.

\subsection{Compositional Strategy}

The Compositional strategy interprets each program operation as a function between abstract states, where each program construct is interpreted by \emph{compositionally} applying a corresponding abstract version of the operation.  This closely aligns with the mathematical, theoretical perspective of abstract interpretation.

For this strategy, as shown on the top left of Figure~\ref{fig:prompts}, we represent the program like a tree to guide LLMs to inductively interpret statements.  
By leveraging the subtree information at each program location, they can perform higher-level operations for locations that depend on previously computed abstract states.
The program locations are not explicit in the abstract program semantics, so we model updating the abstract state at a specific location as a side-effect.
For example, when \texttt{a := read()} is processed in Step 1, we update the abstract state at $\{P1\}$ to be $\{a : [-\inf, \inf]\}$ as a side-effect of interpreting the statement that precedes it.

\begin{figure}[t]
    \centering
    \includegraphics[width=\linewidth, height=12cm]{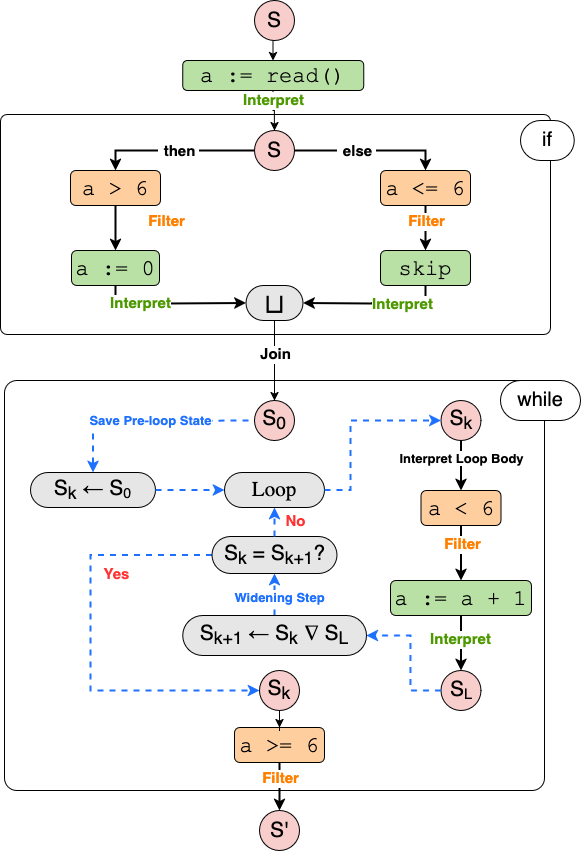}
    \caption{Overall flow of the Compositional strategy for our running example. It corresponds to the high-level workflow shown on the bottom left of Figure~\ref{fig:prompts}.}
    \label{fig:comp-absint}
\end{figure}

Figure~\ref{fig:comp-absint} shows the inner workings of the Compositional strategy.  Black arrows represent the flow between program components, and blue arrows represent the flow within the internal machinery for fixpoint computation. 
This approach takes in an initial abstract state $S$, and iteratively transforms it by processing each statement, and returns a final abstract state $S'$.

Consider the example program in Figure~\ref{fig:annotated_program}. First, we interpret the \texttt{read()} and then go through the \texttt{if}-statement.  The two branches (\texttt{then} and \texttt{else}) are interpreted separately, and their results are joined at the end ($\sqcup$).  
Now, we go through the \texttt{while}-loop, which is interpreted using fixpoint computation in a recursive manner.
We first initialize the iteration at $k=0$, then interpret the loop body, and perform widening ($\nabla$), until we reach $S_k = S_{k+1}$ (a fixpoint).
Upon convergence, we go through filtering again to exit the loop and output our final abstract state $S'$.

\subsection{Transitional Strategy}

\begin{figure}[t]
    \centering
    {\small
    {\setlength{\jot}{0pt}
    \begin{align*}
        M(\{P0\}) &= \{a : [-\inf, \inf]\} \\
        M(\{P1\}) &= Interpret(a := read, P0(a)) \\
        M(\{P2\}) &= Filter(a > 6, P1(a)) \\
        M(\{P3\}) &= Interpret(a := 0, P2(a)) \\
        M(\{P4\}) &= Filter(a \leq 6, P1(a)) \\
        M(\{P5\}) &= Interpret(skip, P4(a)) \\
        M(\{P6\}) &= \{a : P3(a) \sqcup P5(a)\} \\
        M(\{P7\}) &= Filter(a < 6, P6(a) \sqcup P8(a)) \\
        M(\{P8\}) &= Interpret(a := a+1, P7(a)) \\
        M(\{P9\}) &= Filter(a \geq 6, P6(a) \sqcup P8(a))
    \end{align*}
    }
    }
    \vspace{-20pt}
    \caption{Fixed point equations for Transitional Prompting for our running example.}
    \label{fig:fixpoint_equations}
\end{figure}

In contrast to Compositional strategy, which inductively reasons over program statements, the Transitional strategy explicitly derives and solves a system of \emph{fixpoint equations} (FPE).  FPEs capture how abstract states are transformed based on the program semantics and control flow. Each program location has a corresponding FPE, and the system of equations is solved using a standard worklist algorithm.  This closely aligns with how abstract interpreters are implemented in practice, known as chaotic iterations \cite{DBLP:conf/ershov/Bourdoncle93}. 

For this strategy, as shown on the right-hand side of Figure~\ref{fig:prompts}, we first ask LLMs to come up with a set of FPEs.
Figure~\ref{fig:fixpoint_equations} shows the set of FPEs for our running example.  
For instance, the abstract state at $\{P7\}$ (loop head) is the result of filtering the join of the abstract states at $\{P6\}$ (before the loop) and $\{P8\}$ (after the loop body) by the loop guard ($\mathtt{a < 6}$).

Then, we ask the models to solve it in a linear fashion using a worklist algorithm. 
The worklist is a list of program locations whose abstract states have not yet converged.
Initially, the worklist contains all program locations, and the procedure continues until the worklist is completely empty.

For every program location $\{P_x\}$ that is picked at each step: (1) $\{P_x\}$ is removed from the worklist. (2) The abstract state for $\{P_x\}$, $S$, is calculated based on its FPE.  (3a) $S$ is saved to a map $M$, where $M(\{P_x\})$ is the most recent abstract state for $\{P_x\}$.  (3b) If $\{P_x\}$ is the first program location inside of a while-loop body, the most recent abstract state for $\{P_x\}$ (i.e., $M(\{P_x\})$) is widened by the current abstract state $S$ to ensure termination.  (4) If $M(\{P_x\})$ has changed during the update, then the program locations whose FPEs directly depend on $\{P_x\}$ are added to the worklist.  This procedure continues until the worklist is finally empty.

\section{Evaluation and Results}
\label{section:evaluation}

\label{sec:raw-scores}

\begin{table*}
    \small
    \centering
    \caption{Comparison of different models across the two strategies on 22 C programs.}
    \label{table:main-raw-scores}
    \resizebox{\textwidth}{!}{
    \begin{tabular}{l cccc cccccccc}
        \toprule
        \multirow{3}{*}{\textbf{Program}}
        & \multicolumn{4}{c}{\textbf{Compositional}} 
        & \multicolumn{8}{c}{\textbf{Transitional}} \\

        \cmidrule(lr){2-5} 
        \cmidrule(lr){6-13} 
        
        & \multicolumn{4}{c}{\textit{Invariant Map Soundness}} 
        & \multicolumn{4}{c}{\textit{Invariant Map Soundness}} 
        & \multicolumn{4}{c}{\textit{Fixpoint Equation Correctness}} \\

        & Llama & Gemini & GPT-4o & QwQ
        & Llama & Gemini & GPT-4o & QwQ 
        & Llama & Gemini & GPT-4o & QwQ \\
        \midrule
        afnp2014.c                  
        & $3/7$ & $7/7$ & $6/7$ & $7/7$
        & $3/7$ & $7/7$ & $3/7$  &  -
        & $7/7$ & $7/7$ & $7/7$ & $7/7$ \\
        as2013-hybrid.c             
        & $3/14$ & $14/14$ & $14/14$ &  $14/14$
        & $11/14$ & - & $3/14$  &  -
        & $12/14$ & $14/14$ & $14/14$ & $13/13$ \\
        benchmark02\_linear.c       
        & $11/12$ & $12/12$ & $12/12$  &  $12/12$
        & $9/12$ & $12/12$ & $12/12$ &  $12/12$
        & $10/12$ & $12/12$ & $12/12$ & $12/12$ \\
        benchmark04\_conjunctive.c  
        & $12/13$ & $13/13$ & $6/13$ &  $2/13$
        & $9/13$ & $13/13$ & $6/13$ &  -
        & $11/13$ & $13/13$ & $12/13$  & $13/13$ \\
        cggmp2005.c                 
        & $5/9$ & $9/9$ & $8/9$ &  $6/9$
        & $3/9$ & $8/9$ & - &  -
        & $9/9$ & $9/9$ & $9/9$ & $9/9$ \\
        const.c                     
        & $14/14$ & $14/14$ & $14/14$ &  $14/14$
        & $14/14$ & $14/14$ & $14/14$ &  $14/14$
        & $14/14$ & $14/14$ & $14/14$ & $14/14$ \\
        count\_by\_2.c              
        & $6/6$ & $6/6$ & $6/6$ &  $5/6$
        & $4/6$ & $6/6$ & $6/6$ &  $5/6$
        & $6/6$ & $6/6$ & $6/6$ & $2/6$ \\
        css2003.c                   
        & $10/16$ & $16/16$ & $13/16$ &  $14/16$
        & $8/16$ & $16/16$ & - &  -
        & $14/16$ & $16/16$ & $16/16$ & $16/16$ \\
        deep-nested.c               
        & $10/33$ & - & $10/33$ & $9/33$
        & $4/33$ & $21/33$ & - &  -
        & $33/33$ & $33/33$ & $33/33$ & $13/33$ \\
        eq1.c                       
        & $14/14$ & $14/14$ & $14/14$ &  $14/14$
        & $14/14$ & $14/14$ & $14/14$ & $0/14$
        & $14/14$ & $14/14$ & $14/14$ & $5/14$ \\
        eq2.c                       
        & $9/9$ & $9/9$ & $9/9$ &  $9/9$
        & $9/9$ & $9/9$ & $9/9$ &  $5/9$
        & $9/9$ & $9/9$ & $9/9$ & $2/9$ \\
        even.c                      
        & $5/5$ & $5/5$ & $5/5$ &  $5/5$
        & $5/5$ & $5/5$ & $5/5$ &  $5/5$
        & $5/5$ & $5/5$ & $5/5$ & $3/5$ \\
        gauss\_sum.c                
        & $9/14$ & $14/14$ & $13/14$ &  $13/14$
        & $10/14$ & $14/14$ &  $10/14$ &  -
        & $14/14$ & $14/14$ & $14/14$ & $14/14$ \\
        in-de20.c                   
        & - & $14/14$ & $14/14$ &  $8/14$
        & $7/14$ & $13/14$ & $5/14$ &  $10/14$
        & $14/14$ & $14/14$ & $14/14$ & $10/14$ \\
        jm2006.c                    
        & $13/18$ & $18/18$ & $16/18$ &  $15/18$
        & $6/18$ & $18/18$ & - &  $11/18$
        & $18/18$ & $18/18$ & $17/18$ &  $18/18$\\
        loopv3.c                    
        & $8/11$ & $11/11$ & $11/11$ &  $5/11$
        & $9/11$ & $11/11$ & $11/11$ &  $9/11$
        & $11/11$ & $11/11$ & $11/11$ & $6/11$ \\
        mine-2018-ex4.6.c           
        & $5/5$ & $5/5$ & $2/5$ &  $5/5$
        & $5/5$ & $5/5$ & $3/5$ &  $5/5$
        & $5/5$ & $5/5$ & $3/5$ & $5/5$ \\
        mono-crafted\_7.c           
        & $6/17$ & $14/17$ & $13/17$ &  $7/17$
        & $13/17$ & - & - &  -
        & $17/17$ & $17/17$ & $17/17$ & $17/17$ \\
        Mono6\_1.c                  
        & $4/12$ & $12/12$ & $12/12$ &  $12/12$
        & $7/12$ & - & $6/12$ &  -
        & $12/12$ & $12/12$ & $12/12$ & $12/12$ \\
        nested\_1.c                 
        & $6/11$ & $11/11$ & $11/11$ &  $11/11$
        & $11/11$ & $11/11$ & $11/11$ &  $11/11$
        & $11/11$ & $11/11$ & $11/11$ & $11/11$ \\
        nested\_2.c                 
        & $10/16$ & $16/16$ & $16/16$ &  $5/16$
        & $15/16$ & $15/16$ & $10/16$ &  -
        & $16/16$ & $16/16$ & $16/16$ & $13/16$ \\
        simple\_vardp\_1.c          
        & $9/9$ & $9/9$ & $9/9$ &  -
        & $4/9$ & $9/9$ & $8/9$ &  -
        & $9/9$ & $9/9$ & $9/9$ & $9/9$ \\
        \bottomrule
    \end{tabular}
    }
\end{table*}

\subsection{Experimental Setup}

\paragraph{Implementation Details} 
We selected 22 C programs from the SV-COMP 2019 dataset~\citep{beyer2019svcomp} containing complex control flows, such as nested loops and conditionals. C programs were parsed to IMP using a customized parser.
Once the models are queried, we automatically verify the soundness of the invariant map using UAutomizer~\citep{10.1007/978-3-642-36742-7_53}, a winning tool in the latest SV-COMP.

\paragraph{Models} For our main experiment, we selected four models: 
(1) \emph{Llama}~\citep{nvidia}: NVIDIA's Llama 3.1 Nemotron 70B Instruct, 
(2) \emph{Gemini}~\citep{gemini2}: Google's Gemini 2.0 Flash,
(3) \emph{GPT-4o}~\citep{openai2024gpt4technicalreport}: OpenAI's GPT-4o,
and (4) \emph{QwQ}~\citep{qwen2}: Qwen's QwQ 32B Preview.
All queries were made using their native API libraries, except for Llama which used OpenRouter.\footnote{https://openrouter.ai/} We set the temperature to 0 across models for stability.

\paragraph{Research Questions} Our experimental evaluation is motivated by the following research questions:

\begin{enumerate}
    \item[\textbf{RQ 1:}] Can LLMs generate sound invariants under our prompting strategies?
    \item[\textbf{RQ 2:}] Does the strategy (Compositional or Transitional) have an impact on the generation of the invariants and the correctness of the reasoning steps?
    \item[\textbf{RQ 3:}] Can LLMs generate sound reasoning traces during invariant generation in the style of abstract interpreters guided by our prompting strategies?
\end{enumerate}

\subsection{Main Numerical Results} 

A key metric used to measure the correctness of the LLMs is the number of program locations where the invariant map was sound. Our main numerical results are presented in Table~\ref{table:main-raw-scores}, comparing the ability to generate sound invariants when prompted with the Compositional and the Transitional strategies.  For the \emph{Invariant Map Soundness} columns, the fractions correspond to the percentage of program locations for which a sound invariant map was generated.  `$-$' indicates that no valid invariant maps were returned. For the Transitional strategy, each entry in the \emph{Fixpoint Equation Correctness} column represents the percentage of program locations for which a sound fixpoint equation was generated. The results answer RQ1 affirmatively, showing that the LLMs have the ability to generate and return sound invariants.

\paragraph{Compositional Results}
Table~\ref{table:main-raw-scores} demonstrates that all models are relatively successful in generating sound invariant maps.  
There were only three instances where an invariant map was not returned across the models.  The first case is (Llama, \texttt{in-de20.c}), where the reasoning process did not begin and was cut-off preemptively; we do not have any hypotheses as to why it occurred for this program, other than the stochastic nature of language models.  In the case of (Gemini, \texttt{deep-nested.c}), the reasoning process terminated prematurely during fixpoint computation; this is relatively unsurprising due to \texttt{deep-nested.c} having more than five nested loops.  Lastly, in the case of (QwQ, \texttt{simple\_vardp\_1.c}), it appears that the model attempts to output the final abstract state, but is cut-off during the reasoning process.

\paragraph{Transitional Results}
In the case of the Transitional Strategy, all models generally seemed to perform better at generating the fixpoint equations compared to solving them.  Surprisingly, Llama was able to return a final invariant map, whereas Gemini, GPT-4o, and QwQ were unable to do so in many cases.  Upon manual inspection, every time `$-$' was returned, the model did not complete an attempted fixpoint computation, with the exception of (GPT-4o, deep-nested.c).  In this case, the model acknowledged the complexity of the program and the large number of nested loops, and gave up.  For Llama, the model omitted the majority of reasoning steps and just returned a final invariant map.  This could indicate that Llama is inferring invariants based on the program syntax rather than reasoning about it formally.  While unsuccessful, the other models appear to make an effort to derive the invariant map through abstract interpretation and show each step of their reasoning.

\begin{table}[h]
\centering
\caption{Invariant Map Soundness Differences (Transitional - Compositional)}
\label{table:invariant-map-differences}
 \resizebox{\columnwidth}{!}{
\begin{tabular}{l cccc}
\toprule
\textbf{Program} & \textbf{Llama} & \textbf{Gemini} & \textbf{GPT-4o} & \textbf{QwQ} \\
\midrule
afnp2014.c & $0/7$ & $0/7$ & $-3/7$ & - \\
as2013-hybrid.c & $8/14$ & - & $-11/14$ & - \\
benchmark02\_linear.c & $-2/12$ & $0/12$ & $0/12$ & $0/12$ \\
benchmark04\_conjunctive.c & $-3/13$ & $0/13$ & $0/13$ & - \\
cggmp2005.c & $-2/9$ & $-1/9$ & - & - \\
const.c & $0/14$ & $0/14$ & $0/14$ & $0/14$ \\
count\_by\_2.c & $-2/6$ & $0/6$ & $0/6$ & $0/6$ \\
css2003.c & $-2/16$ & $0/16$ & - & - \\
deep-nested.c & $-6/33$ & - & - & - \\
eq1.c & $0/14$ & $0/14$ & $0/14$ & $-14/14$ \\
eq2.c & $0/9$ & $0/9$ & $0/9$ & $-4/9$ \\
even.c & $0/5$ & $0/5$ & $0/5$ & $0/5$ \\
gauss\_sum.c & $1/14$ & $0/14$ & $-3/14$ & - \\
in-de20.c & - & $-1/14$ & $-9/14$ & $2/14$ \\
jm2006.c & $-7/18$ & $0/18$ & - & $-4/18$ \\
loopv3.c & $1/11$ & $0/11$ & $0/11$ & $4/11$ \\
mine-2018-ex4.6.c & $0/5$ & $0/5$ & $1/5$ & $0/5$ \\
mono-crafted\_7.c & $7/17$ & - & - & - \\
Mono6\_1.c & $3/12$ & - & $-6/12$ & - \\
nested\_1.c & $5/11$ & $0/11$ & $0/11$ & $0/11$ \\
nested\_2.c & $5/16$ & $-1/16$ & $-6/16$ & - \\
simple\_vardp\_1.c & $-5/9$ & $0/9$ & $-1/9$ & - \\
\bottomrule
\end{tabular}
}
\end{table}

\begin{table}
    \footnotesize
    \centering
    \caption{Comparison of advanced models on selected C programs. GPT refers to GPT-o1, and DS refers to DeepSeek-R1.}
    \label{table:hard_programs}
    \resizebox{\linewidth}{!}{
    \begin{tabular}{l cc cccc}
        \toprule
        \multirow{3}{*}{\textbf{Program}}
        & \multicolumn{2}{c}{\textbf{Compositional}} 
        & \multicolumn{4}{c}{\textbf{FPE}} \\
        
        \cmidrule(lr){2-3} \cmidrule(lr){4-7}
        & \multicolumn{2}{c}{\textit{IM Soundness}} 
        & \multicolumn{2}{c}{\textit{IM Soundness}} 
        & \multicolumn{2}{c}{\textit{FPE Correctness}} \\

        & GPT & DS 
        & GPT & DS 
        & GPT & DS \\
        \midrule
        deep-nested.c     & $10/33$ & $8/33$ & $8/33$ & $31/33$ & $31/33$ & $33/33$ \\
        mono-crafted\_7.c & $17/17$ & $17/17$ & $17/17$ & $17/17$ & $17/17$ & $14/17$ \\
        nested\_1.c       & $11/11$ & $11/11$ & $11/11$ & $11/11$ & $11/11$ & $11/11$ \\
        nested\_2.c       & $16/16$ & $16/16$ & $16/16$ & $16/16$ & $16/16$ & $13/16$ \\
        \bottomrule
    \end{tabular}
    }
\end{table}

\paragraph{Comparing the Two Strategies}
To better understand the differences in performance between the two strategies, we consider their quantitative differences.
Table~\ref{table:invariant-map-differences} computes the difference in the \emph{Invariant Map Soundness} scores.  If final invariant maps were returned for both strategies for a given model, the entry corresponding to the model indicates the difference between the \emph{Invariant Map Soundness} scores.  Positives scores indicate that for that particular program, the model performed better using the Transitional strategy, and vice versa.  It can be seen that Llama, GPT-4o, and QwQ may have dramatic differences between the performances of the two strategies.  For instance, QwQ has a difference score of $-14$ for \texttt{eq\_1.c}.  On the other hand, Gemini has relatively consistent scores when both strategies are able to elicit a final invariant map,  with the largest difference being $1$, in favor of the transitional strategy.  These results suggest the possibility that certain models, when evaluated on certain programs, take on reasoning styles more suitable to one strategy, which answers our RQ2.

Across both of the strategies, the models struggle with \texttt{deep-nested.c} in terms of generating sound invariant maps.  This is relatively unsurprising, as \texttt{deep-nested.c} is a program with much more complex control flow compared to the other benchmark programs.  This begs the question of whether the reasoning capability of the models could impact their performance.  Therefore, we tested both strategies on two models that were specifically trained for complex reasoning tasks, Deepseek-R1~\citep{deepseekai2025deepseekr1incentivizingreasoningcapability} and GPT-o1~\citep{openai2024gpt4o}.  As a reference point, we also tested both models on other programs with complex control flow (e.g., nested loops): \{\texttt{deep-nested.c}, \texttt{mono-crafted\_7.c}, \texttt{nested\_1.c}, \texttt{nested\_2.c}\}.  These \\
results are displayed in Table~\ref{table:hard_programs}.  For \texttt{deep-nested.c}, GPT-o1 struggles to generate sound invariants across both strategies.  Deepseek seems to perform better at generating a sound invariant map for the Transitional strategy, but like GPT-o1, struggles with the Compositional Strategy.  On the remaining programs, GPT-o1 and Deepseek have the same results in terms of soundness.  However, Deepseek generated several unsound FPEs for \texttt{mono-crafted\_7.c} and \texttt{nested\_2.c}. 

Despite reasonable performance scores, these metrics only capture the end result, not the reasoning process that led to it. 
In the context of invariant generation, it remains unclear whether LLMs arrive at correct answers through sound logical derivations or through heuristics and pattern-matching based on the program. 
Our prompting strategies, which elicit step-by-step explanations from LLMs, allow us to thematically analyze their outputs and assess the extent to which they emulate formal reasoning. 
Unlike traditional approaches in LLM-aided invariant generation, our analysis provides a complementary perspective by investigating how LLMs arrive at invariants, which helps to reveal both their capabilities and limitations in producing reliable invariants.

\section{Key Thematic Errors and Possible Opportunities}
\label{section:error_analysis}

\lstset{
  basicstyle=\footnotesize\ttfamily
}

In this section, we describe the key types of mistakes the LLMs made in the reasoning process across both strategies.  This answers RQ3, showing the LLM's reasoning traces and errors made during the process.

\subsection{Understanding Control Flow}

Understanding the control flow of a program is a core competency required to do abstract interpretation effectively.  This entails understanding how abstract states impact others, following how abstract states propagate throughout the program.

\subsubsection{Compositional Strategy}
\label{sec:cs6}
Under the Compositional strategy, a good understanding of control flow allows for correctly reasoning about the program inductively based on its structure.
Here, we highlight a few cases where the abstract states were not correctly calculated inductively.  It seems that, in general, the models were capable of correctly identifying the structures of a program (e.g., what the statements are, the high level order of the statements, and how to compose them), but had several key issues regarding keeping track of \emph{abstract} control flow. 

Sometimes, models had issues propagating the abstract states correctly. Consider the program snippet in Listing~\ref{lst:lstcountby2}.

\begin{lstlisting}[caption={count\_by\_2.c Snippet}, captionpos=b, xleftmargin=.1\linewidth, xrightmargin=.1\linewidth, backgroundcolor=\color{blue!7},label={lst:lstcountby2}]
@B@{P0}@BE@
/*@$i := \mathtt{read()}$;@*/
@B@{P1}@BE@
/*@$i := 0$;@*/
@B@{P2}@BE@
...
\end{lstlisting}
%
While analyzing the program, QwQ initially correctly infers that at \textcolor{blue}{\{P1\}}, the abstract state for $i$ is $[-\inf, \inf]$, but after interpreting $i := 0$, when the abstract state for $i$ should become $[0, 0]$ at \textcolor{blue}{\{P2\}}, the model becomes confused and claims that this line has an effect on \textcolor{blue}{\{P1\}}.  
While \textcolor{blue}{\{P2\}}'s abstract state should have no impact on \textcolor{blue}{\{P1\}}'s abstract state, it incorrectly propagates \textcolor{blue}{\{P2\}}'s abstract state to \textcolor{blue}{\{P1\}}.

Other times, models did not cover all the possible paths through which abstract information can flow.  Consider the code snippet in Listing~\ref{lst:loopv3v1}.

\begin{lstlisting}[caption={loopv3.c Snippet 1}, captionpos=b, xleftmargin=.1\linewidth, xrightmargin=.1\linewidth, backgroundcolor=\color{blue!7},
label={lst:loopv3v1}]
...
@B@{P2}@BE@
while (i < 50000001) do
  @R@[while_true]@RE@
  @B@{P3}@BE@
  ...
  @B@{P8}@BE@
end @R@[while_false]@RE@
...
\end{lstlisting}
%
For this code snippet, QwQ does not properly account for all of \textcolor{blue}{\{P3\}}'s dependencies.  Specifically, during fixpoint computation, it forgets to account for the fact that an abstract state should flow from \textcolor{blue}{\{P2\}} to \textcolor{blue}{\{P3\}}, leading to unsoundness in the abstract state for \textcolor{blue}{\{P3\}}. 

\subsubsection{Transitional Strategy}
Under the Transitional strategy, a good understanding of control flow corresponds to the correct generation of fixpoint equations (FPEs), since they are a direct (but abstract) reflection of data flows through the program.
Here, we highlight a few cases where the fixpoint equations were incorrect. 

Sometimes, models appeared to not grasp the concept of what FPEs are (e.g., GPT-4o for \texttt{mine-2018-ex4.6.c}).  Instead of writing FPEs as expected, LLMs assigned concrete intervals to each program location based on an unsound analysis that completely ignored loop structure.  

Sometimes, models produced incorrect FPEs because they did not know how to properly formulate abstract joins for a loop (e.g., Llama for \texttt{as-2013-hybrid.c}, \texttt{benchmark02\_\\linear.c}, and \texttt{benchmark04\_conjunctive.c}).  
For instance, consider the code snippet in Listing~\ref{lst:as2013snippet}.
The FPE for location \textcolor{blue}{$\{P5\}$} is incorrectly written as $M(\{P5\})\\ = Filter(j < 10, M({P4}) \sqcup M({P7}))$, when it should be $M(\{P5\}) \\= Filter(j < 10, M({P4}) \sqcup M({P6}))$.  Similar errors occurred for the other programs.

 \begin{lstlisting}[caption={as2013-hybrid.c Snippet}, captionpos=b, xleftmargin=.1\linewidth, xrightmargin=.1\linewidth, backgroundcolor=\color{blue!7},
 label={lst:as2013snippet}]
...
@B@{P4}@BE@
while (j < 10) do
  @R@[while_true]@RE@
  @B@{P5}@BE@
  /*@$j := j + 1$;@*/
  @B@{P6}@BE@
end @R@[while_false]@RE@
@B@{P7}@BE@
...
\end{lstlisting}

Finally, LLMs sometimes yielded FPEs where the control flow was shifted, meaning that the abstract states were propagated incorrectly (e.g., QwQ for \texttt{count\_by\_2.c}, \texttt{eq2.c}, \texttt{in-de20.c}, \texttt{loopv3.c}).
Consider the code snippet in Listing~\ref{lst:loopv3v2}.
\begin{lstlisting}[caption={loopv3.c Snippet 2}, captionpos=b, xleftmargin=.1\linewidth, xrightmargin=.1\linewidth, backgroundcolor=\color{blue!7},
label={lst:loopv3v2}]
...
@B@{P3}@BE@
if (!(read() == 0)) then
  @R@[if_then]@RE@
  @B@{P4}@BE@
  /*@$i := i + 8$;@*/
  @B@{P5}@BE@
...
\end{lstlisting}

The FPEs for \textcolor{blue}{$\{P4\}$} and \textcolor{blue}{$\{P5\}$} returned by QwQ were \\
$M(\{P4\}) = \mathtt{Interpret}(i := i + 8, \mathtt{Filter}(!(\mathtt{read()} == 0), \\ M(\{P3\}))$ and $M(\{P5\}) = M(\{P4\})$. This is incorrect, as \textcolor{blue}{$\{P4\}$} represents the abstract state prior to interpreting $i := i + 8;$.

\subsection{Fixpoint Computation}

Proper fixpoint computation is critical to the soundness of an abstract interpreter.  In this subsection, we describe some of the key errors in fixpoint computation we observed.

A common error we found was incorrectly interpreting widening, which may terminate the fixpoint computation prematurely.  Consider the following subset of the reasoning trace when GPT-4o analyzes the \texttt{mono\_crafted\_7.c} program, using the Compositional Strategy in Listing~\ref{lst:mono7reason}.

\begin{lstlisting}[caption={mono\_crafted\_7.c Reasoning Trace}, captionpos=b, backgroundcolor=\color{red!7},
label={lst:mono7reason}]
The input abstract state to this 
iteration is
/*@$\{x : [1000000, \inf], y : [50000, \inf], z : [0, 0]\}$@*/
...
/*@${x : [1000000, \inf], y : [50000, \inf], z : [0, 0]} \nabla$@*/
/*@${x : [999998, \inf], y : [49998, \inf], z : [0, 0]}$ results in@*/
/*@${x : [1000000, \inf], y : [50000, \inf], z : [0, 0]}$@*/
We are at a fixed point.
\end{lstlisting}

Fixpoint computation terminates prematurely in this reasoning trace because of a widening operation error.
Widening $[1000000, \inf]$ by $[999998, \inf]$ should result in $[-\inf, \inf]$, rather than $[1000000, \inf]$.  This is not just imprecise, but is in fact, unsound. 

Another key issue is detecting when a fixpoint has been reached.  Consider the following subset of the reasoning trace when Gemini analyzes the program \texttt{as2013\_hybrid.c}, using the Transitional Strategy in Listing~\ref{lst:as13reason}.

\begin{lstlisting}[caption={as2013\_hybrid.c Reasoning Trace}, captionpos=b, backgroundcolor=\color{red!7},
label={lst:as13reason}]
* Compute F_3(M), and 
  update /*@M({P3}) = ${i : [1, \inf], j : [-\inf, \inf]}$@*/.
* Add {P4} to W.
...
* Compute F_3(M), and 
  update /*@M({P3}) = ${i : [1, \inf], j : [-\inf, \inf]}$@*/.
* Add {P4} to W.
\end{lstlisting}

We can see that throughout the chaotic iterations, the abstract state at \textcolor{blue}{\{P3\}} does not change, indicating that a fixpoint has been reached for \textcolor{blue}{\{P3\}}.  However, Gemini does not seem to detect this and unnecessarily adds \textcolor{blue}{\{P4\}} (which depends on \textcolor{blue}{\{P3\}}) to the worklist again.

The \texttt{loopv3.c} snippet in Listing~\ref{lst:loopv3v1} (using the Compositional strategy with the QwQ model) is an example of when fixpoint iteration is conducted improperly, due to the lack of understanding the control flow.  The abstract state from \textcolor{blue}{\{P2\}} is not incorporated as a result of not joining (or widening with) the result from the previous fixpoint iteration. This error propagates, leading to unsound reasoning further down the chain of reasoning.

\subsection{Operation-Based Errors}
\label{subsec:op_err}
We observe that LLMs may overlook or misinterpret essential program operations that are crucial in ensuring sound abstract states.  
For instance, in \texttt{as2013-hybrid.c}, Gemini under the Compositional strategy incorrectly filters $[0, 0]$ with $i \leq 9$ to $[0, 9]$, overapproximating unintentionally when it already satisfies the condition.
Under the Transitional strategy, LLMs often neglected the widening operator during worklist iterations, especially in complex nested structures requiring prolonged analysis (e.g., GPT-4o for for \texttt{css2003.c}).

Consider the subset of the reasoning trace when QwQ analyzes the program \texttt{mono\_crafted\_7.c}, using the Compositional strategy in Listing~\ref{lst:mono7reason2}.
QwQ first makes a glaring error in calculating for the widening operation, where the results should have been in $x : [-\inf, \inf]$ and $y: [-\inf, 50000]$. The model makes such a mistake because it ignores that the operation also works on $x$.  Focused on analyzing the range of $y$ with its loop condition, QwQ overlooks the operations on $x$, gets confused by its context, and fails to analyze its correct result. 

\subsection{Short-Circuiting}
Short-circuiting occurs when the model generates a logical step in its reasoning trace that does not align with the actual computations and appears to skip multiple intermediate
\begin{lstlisting}[caption={mono\_crafted\_7.c Reasoning Trace}, captionpos=b,
  backgroundcolor=\color{red!7},
  basicstyle=\ttfamily\footnotesize, columns=fullflexible,
  breaklines=true, breakatwhitespace=false,
  label={lst:mono7reason2}]
Widening with previous iteration input:
{x: [1000000, inf], y: [50000,50000], z: [0,0]} nabla
{x: [999998, inf], y: [49998,49998], z: [0,0]} =
{x: [999998, inf], y: [49998, inf],  z: [0,0]}
...
Looking at the widening operation, each iteration lowers x and y by 2,
but widening keeps track of the lower bound decreasing.
In practice, after a number of iterations, the state will stabilize.
Since y is decreasing by 2 each time and we're widening, the lower bound
of y keeps dropping; eventually y <= 0.
But to simplify, perhaps after some iterations y reaches [0, inf],
and then the loop guard y > 0 filters to y in (0, inf];
but since y is decremented by 2 each time, it keeps decreasing.
This seems tricky.
...
So x is in [950000, inf], y is [0,0]
\end{lstlisting}
steps. Using the same subset of the reasoning trace in Listing~\ref{lst:mono7reason2}, where QwQ analyzes the program \texttt{mono\_crafted\_7.c} \\ using the Compositional Strategy, we observe that the reasoning following the widening operation error exhibits precisely this short-circuiting behavior.

In its futile attempt at self-correction of the incorrect widening operation, QwQ resorts to ambiguous natural language to describe updates to the abstract states, without specifying the concrete operations involved, which is required in abstract formal reasoning.  The only instance where an operator is mentioned is during the discussion of the loop guard and filtering. However, this is ultimately irrelevant to the final unsupported conclusion that “$y$ is $[0,0]$.”

\subsection{Limitations of LLMs}
While our evaluation focuses on how LLMs reason about abstract interpretation, we observed certain limitations that stem not from the difficulty of the task itself, but from inherent limitations in the LLMs.

One common issue was the context window limit, which degraded LLMs' performance, as shown in QwQ's example in Section~\ref{subsec:op_err}.  
According to Qwen2's technical report~\cite{qwen2}, it is likely that LLM will have less attention on the the middle part of the prompt, and this will lead to such kinds of errors occurring. 
This is also confirmed in our experiments, where models often began strong, producing accurate abstract states and sound reasoning steps in the early stages of analysis.  But their performance gradually degraded, exhibiting signs of ``forgetting'' prior context as they progressed.  This led to logically inconsistent or incomplete invariants in later program locations, even though similar reasoning patterns had been successfully applied earlier in the analysis.

Another common issue was premature output truncation caused by output token limits.  For programs with complex control-flow structures (e.g., deeply nested conditionals or loops with many iterations), LLMs often failed to complete the fixpoint computation or reach the final abstract state outputs, before their responses were cut off mid-execution.

These issues highlight that some LLM errors are not due to flaws in their reasoning abilities per se, but rather due to practical deployment constraints.  Such findings suggest the need for more stateful approaches or more modular context management to mitigate these issues, especially for tasks like abstract interpretation-based program analysis, where correctness depends on maintaining logical consistency.

\section{Conclusion}
\label{section:conclusion}

In this work, we investigate the ability of LLMs to reason as abstract interpreters.
While recent work has shown that LLMs generate many valid invariants with fine-tuning or verifier feedback, we demonstrate that LLMs have key failures when conducting the analysis themselves out-of-the-box.
We introduce two prompting strategies which generate reasoning traces that can be used as proxies for understanding LLMs' internal reasoning.
Our results and analysis show that LLMs are limited in many aspects, where they make critical errors in understanding control-flow, evaluating operations, and other failures.
We hope that this work serves as a starting point for future research investigations which address these concerns and improve these aspects by introducing novel language model architectures suitable for static program analysis and invariant generation.

\begin{acks}
This research was supported in part by the U.S. National Science Foundation (NSF) under grant CCF-2220345. We thank the anonymous reviewers for their constructive feedback.
\end{acks}


\bibliographystyle{ACM-Reference-Format}
\bibliography{sample-base}


\onecolumn
\appendix
\section{Prompts Used}
\label{appendix:prompts}
\subsection{Compositional Strategy}
\label{appendix:compositional}
\begin{tcolorbox}[breakable, colback=gray!20, colframe=black]
\textbf{Context:} Given a program, analyze the program with abstract interpretation, using the interval abstract domain.
Programs are composed of assignment, skip, if-then-else, while-loops, and sequential composition of these statements,
where program variables are integer variables. The goal is to output an abstract state for each program location.  
An abstract state maps each program variable to an interval, or the empty interval $\bot$. For example, 
$\{x : [1, 4], y : [-1, 3]\}$ means that $x$ can take on values between $1$ and $4$ and $y$ can take on values between $-1$ and $3$.
$\bot$ means that the variable cannot have any concrete value. \\

Each abstract state should be sound.  For instance, if the abstract state at location $\{P\}$ maps $x$ to $[4, 10]$, then in
any concrete execution of the program, the value of $x$ should be between $4$ and $10$ at location $\{P\}$. \\

Arithmetic expressions are interpreted with interval arithmetic.  Be cautious of edge cases in interpreting division
with interval arithmetic.  For example, $[1, 3] / [0, 0] = \bot$, as no valid value results from a division by $0$. 
Furthermore, $[1, 3] / [-2, 3] = [-\inf, \inf]$, as division by 0 may or may not occur. \\

$\mathtt{read()}$ expressions are interpreted as $[-\inf, \inf]$, as reading from the standard input can result in any value. \\

The abstract state at $\{P0\}$, the program entry point, maps each program variable to
$[-\inf, \inf]$, indicating that at the beginning of the program, the variables can have any integer value. \\ 

You should abstractly interpret programs in a denotational style. This means that each program statement is 
interpreted as a function, mapping abstract states to abstract states, 
and we iteratively interpret each statement on an input abstract state.  As the
program is being interpreted, we save the abstract state at a program location after interpreting the statement
preceding it, as a side-effect of the interpretation process. \\

There are several directives in the annotated programs that help keep track of control flow.\\
$\mathtt{[if\_then]}$ means that the input abstract state to the if-statement is filtered to account for the fact that the
guard of the if-statement should hold. \\
$\mathtt{[if\_else]}$ means that the input abstract state to the if-statement is filtered to account for the fact that the
negation of the guard of the if-statement should hold.\\
$\mathtt{[endif]}$ means the the result of interpreting the then-branch on the input abstract state and the result of interpreting
the else-branch on the input abstract state are merged.\\

$\mathtt{[while\_true]}$ means that the input abstract state to a while-statement is filtered to account for the fact that the
loop guard should hold.\\
$\mathtt{[while_false]}$ means that the abstract state as a result of interpreting the loop body is filtered by the negation of the loop guard,
indicating possible behaviors when the while loop is no longer executed. \\

Some examples of filtering are:

\begin{itemize}
    \item[-] Filtering abstract state ${x : [5, 7], y : [6, 8]}$ by $\mathtt{!(read() == 0)}$ results in the same abstract state, because
    we cannot know for certain if the result of reading from standard input is $0$.
    \item[-] Filtering abstract state $\{x : [5, 10], y : [5, \inf]\}$ by $\mathtt{!(y == 6)}$ results in ${x : [5, 10], y : [5, \inf]}$.  Filtering
    by $\mathtt{!(y == 6)}$ is equivalent to filtering by $y > 6 || y < 6$.  Filtering the abstract state by $y > 6$ results in 
    ${x : [5, 10], y : [7, \inf]}$.  Filtering the abstract state by $y < 6$ results in ${x : [5, 10], y : [5, 5]}$. 
    Joining the resulting abstract states results in ${x : [5, 10], y : [5, \inf]}$.
    \item[-] Filtering abstract state ${x : [5, 9], y : [10, 12]}$ by $\mathtt{y == 16}$ results in $\{x : \bot, y : \bot\}$, as it is impossible for
    y to be $16$.
    \item[-] Filtering the abstract state ${x : [5, 10], y : [4, 9]}$ by $\mathtt{(y <= 8) \&\& (x <= y)}$ results in the 
    filtering the state by $\mathtt{y <= 8}$ and filtering the state by $x <= y$ and then intersecting the resulting states.
    Filtering $\{x : [5, 10], y : [4, 9]\}$ by $y <= 8 $results in$ {x : [5, 10], y : [4, 8]}.$  Filtering
   $ {x : [5, 10], y : [4, 9]}$ by $x <= y $ results in ${x : [5, 9], y : [4, 9]}$.  Intersecting both states 
    results in ${x : [5, 9], y : [4, 8]}$.
\end{itemize}

While loops are interpreted using fixed point computation, by iteratively interpreting the body of the while loop on the
input abstract state.  At each iteration, the result from the previous iteration is widened by the interpretation of the loop
body on the abstract state from the previous iteration.  Widening is used to ensure termination of the fixed point computation.
The widening of two intervals is defined symbolically as $[a, b] \nabla [c, d] = [\text{if  } c < a, \text{ then } -\inf \text{ else } a, \text{ if } d > b \text{ then } \inf \text{ else } b]$.  
For example, $[6, 7] \nabla [9, 10] = [6, \inf]$. Note that $\bot \nabla [c, d] = [c, d]$ and $[a, b] \nabla \bot = [a, b]$.
Finally, when a fixed point is reached, it is filtered by the negation of the loop guard, 
which is the final result of interpreting the loop.

When all commands have been interpreted, the most recent abstract state at each program location is returned.

Here are some examples:

\textbf{Example 1}:

\textbf{Input:} \begin{lstlisting}[basicstyle=\ttfamily]
    {P0} 
    x := read();
    {P1}
    if (x < 3) then 
        [if_then] 
        {P2} 
        x := x - 1; 
        {P3} 
        x := x * 2; 
        {P4} 
    else 
        [if_else] 
        {P5} 
        x := x + 2; 
        {P6}
    end [endif] 
    {P7}
\end{lstlisting}

\textbf{Output}:

Initially, the abstract state at ${P0}$ is ${x : [-\inf, \inf]}$.

Begin interpreting the program.

\begin{itemize}
    \item[1.]  Interpret \texttt{x := read();}
        \begin{itemize}
            \item[-] The input abstract state is $\{x : [-\inf, \inf]\}$
            \item[-] The resulting abstract state is $\{x : [-\inf, \inf]\}$
            \item[-] As a side-effect, the abstract state at $\{P1\}$ is $\{x : [-\inf, \inf]\}$
        \end{itemize}
    \item[2.] Interpret the if-then-else statement.
        \begin{itemize}
            \item[-]  Interpret the then-branch.
                \begin{itemize}
                    \item[1.] The input abstract state is $\{x : [-\inf, \inf]\}$.
                    \item[2.] Filter the input state by x < 3.  The resulting abstract state is $\{x : [-\inf, 2]\}$. As a side-effect,
        the abstract state at $\{P2\}$ is $\{x : [-\inf, 2]\}$.
                    \item[3.]  Interpret x := x - 1;
                        \begin{itemize}
                            \item[-]  The input abstract state is $\{x : [-\inf, 2]\}$.
                            \item[-]  The resulting abstract state is $\{x: [-\inf, 1]\}$.
                            \item[-]  As a side-effect, the abstract state at $\{P4\}$ is $\{x : [-\inf, 2]\}$.
                        \end{itemize}
                \end{itemize}
            \item[-] Interpret the else-branch.
                \begin{itemize}
                    \item[1.] The input abstract state is $\{x : [-\inf, \inf]\}$.
                    \item[2.] Filter the input state by x >= 3. The resulting abstract state is $\{x : [3, \inf]\}$. As a side-effect, the abstract state at $\{P5\}$ is $\{x : [3, \inf]\}$.
                    \item[3.] Interpret x := x + 2;
                        \begin{itemize}
                            \item[-] The input abstract state is $\{x : [3, \inf]\}$.
                            \item[-] The resulting abstract state is $\{x : [5, \inf]\}$.
                            \item[-] As a side-effect the abstract state at $\{P6\}$ is $\{x : [5, \inf]\}$.
                        \end{itemize}
             \end{itemize}
             \item[-] Join the results of interpreting the then and else branch:
                \begin{itemize}
                    \item[-] The output of interpreting the then-branch is $\{x : [-\inf, 2]\}$.
                    \item[-] The output of interpreting the else-branch is $\{x : [5, \inf]\}$.
                    \item[-] The result of joining the two states is $\{x : [-\inf, \inf]\}$. As a side-effect, the abstract state at $\{P7\}$ is $\{x : [-\inf, \inf]\}$.
                \end{itemize}
        \end{itemize}
\end{itemize}

There are no more statements to interpret, and the answer is

\begin{align*}
\{P_0\} & \mapsto \{x : [-\inf, \inf]\} \\
\{P_1\} & \mapsto \{x : [-\inf, \inf]\} \\
\{P_2\} & \mapsto \{x : [-\inf, 2]\} \\
\{P_3\} & \mapsto \{x : [-\inf, 1]\} \\
\{P_4\} & \mapsto \{x : [-\inf, 2]\} \\
\{P_5\} & \mapsto \{x : [3, \inf]\} \\
\{P_6\} & \mapsto \{x : [5, \inf]\} \\
\{P_7\} & \mapsto \{x : [-\inf, \inf]\}
\end{align*}

\textbf{Example 2:}

\textbf{Input:} \begin{lstlisting}[basicstyle=\ttfamily]
{P0}
i := 1;
{P1}
j := 0;
{P2} 
while (i <= 5) do 
    [while_true]
    {P3} 
    j := j + i;
    {P4} 
    i := i + 1; 
    {P5} 
end [while_false]
{P6}
\end{lstlisting}

Output: 

Initially, the abstract state at {P0} is ${i : [-\inf, \inf], j : [-\inf, \inf]}$

\begin{itemize}

\item[1.] Interpret i := 1
    \begin{itemize}
        \item[-] The input abstract state is $\{i : [-\inf, \inf], j : [-\inf, \inf]\}$
        \item[-] The resulting abstract state is $\{i : [1, 1], j : [-\inf, \inf]\}$
        \item[-] As a side-effect, the abstract state at $\{P1\}$ is $\{i : [1, 1], j : [-\inf, \inf]\}$
     \end{itemize}
\item[2.] Interpret j := 0
    \begin{itemize}
        \item[-] The input abstract state is $\{i : [1, 1], j : [-\inf, \inf]\}$
        \item[-] The resulting abstract state is $\{i : [1, 1], j : [0, 0]\}$
        \item[-] As a side-effect, the abstract state at $\{P2\}$ is $\{i : [1, 1], j : [0, 0]\}$.
    \end{itemize}
\item[3.] Interpret the while loop.
    \begin{itemize}
    \item[-] The input abstract state (iteration 0) is $\{i : [1, 1], j : [0, 0]\}$.
    \item[-] Begin fixed point iteration.
    \item[-] Fixed point Iteration 1:
        \begin{itemize}
        \item[-] The input abstract state to this iteration is $\{i : [1, 1], j : [0, 0]\}$
        \item[-] Filtering the state by i <= 5 results in the abstract state $\{i : [1, 1], j : [0, 0]\}$.  
          As a side-effect, the abstract state at $\{P3\}$ is $\{i : [1, 1], j : [0, 0]\}$.
        \item[-] Interpret j := j + i;
            \begin{itemize}
            \item[-] The input abstract state is $\{i : [1, 1], j : [0, 0]\}$
            \item[-] The resulting abstract state is $\{i : [1, 1], j : [1, 1]\}$
            \item[-] As a side-effect the abstract state at $\{P4\}$ is $\{i : [1, 1], j : [1, 1]\}$.
            \end{itemize}
        \item[-] Interpret i := i + 1;
            \begin{itemize}
            \item[-] The input abstract state is $\{i : [1, 1], j : [1, 1]\}$.
            \item[-] The resulting abstract state is $\{i : [2, 2], j : [1, 1]\}$.
            \item[-] As a side-effect the abstract state at $\{P5\}$ is $\{i : [2, 2], j : [1, 1]\}$.
            \end{itemize}
        \item[-] Widen the input abstract state by the interpretation of the loop body
            \begin{itemize}
                \item[-] The input abstact state to this iteration is $\{i : [1, 1], j : [0, 0]\}$
                \item[-] The result of interpreting the loop body is $\{i : [2, 2], j : [1, 1]\}$.
                \item[-] $\{i : [1, 1], j : [0, 0]\} \nabla \{i : [2, 2], j : [1, 1]\}$ results in $\{i : [1, \inf], j : [0, \inf]\}$.
            \end{itemize}
        \item[-] The result of this iteration is $\{i : [1, \inf], j : [0, \inf]\}$.
        \end{itemize}
    \item[-] Fixed point Iteration 2:
        \begin{itemize}
        \item[-] The input abstract state to this iteration is $\{i : [1, \inf], j : [0, \inf]\}$.
        \item[-] Filtering the state by i <= 5 results in the abstract state $\{i : [1, 5], j : [0, \inf]\}$.  As a side-effect,
        the abstract state at $\{P3\}$ is $\{i : [1, 5], j : [0, \inf]\}$.
        \item[-] Interpret j := j + i;
            \begin{itemize}
                \item[-] The input abstract state is $\{i : [1, 5], j : [0, \inf]\}$.
                \item[-] The resulting abstract state is $\{i : [1, 5], j : [1, \inf]\}$
                \item[-] As a side-effect the abstract state at $\{P4\}$ is $\{i : [1, 5], j : [1, \inf]\}$
            \end{itemize}
        \item[-] Interpret i := i + 1;
            \begin{itemize}
                \item[-] The input abstract state is $\{i : [1, 5], j : [1, \inf]\}$
                \item[-] The resulting abstract state is $\{i : [2, 6], j : [1, \inf]\}$
                \item[-] As a side-effect the abstract state at $\{P5\}$ is $\{i : [2, 6], j : [1, \inf]\}$.
            \end{itemize}
        \item[-] Widen the abstract state from the previous iteration by the interpretation of the loop body
            \begin{itemize}
                \item[-] The input abstract state to this iteration is $\{i : [1, \inf], j : [0, \inf]\}$
                \item[-] The result of interpreting the loop body is $\{i : [2, 6], j : [1, \inf]\}$.
                \item[-] $\{i : [1, \inf], j : [0, \inf]\} \nabla \{i : [2, 6], j : [1, \inf]\}$ results in $\{i : [1, \inf], j : [0, \inf]\}$.
            \end{itemize}
        \item[-] The result of this iteration is $\{i : [1, \inf], j : [0, \inf]\}$.
        \end{itemize}
    \item[-] We are at a fixed point.  The result of the iteration was the same as the previous one.
    \item[-] Filter the fixed point by the negation of the loop-guard, i > 5.  Filtering ${i : [1, \inf], j : [0, \inf]}$ by $i > 5$
    results in $\{i : [6, \inf], j : [0, \inf]\}$.  As a side effect the abstract state at $\{P6\}$ is $\{i : [6, \inf], j : [0, \inf]\}$.
    \end{itemize}
\end{itemize}

There are no more statements to interpret, and the answer is

\begin{align*}
\{P_0\} & \mapsto \{i : [-\inf, \inf], j : [-\inf, \inf]\} \\
\{P_1\} & \mapsto \{i : [1, 1], j : [-\inf, \inf]\} \\
\{P_2\} & \mapsto \{i : [1, 1], j : [0, 0]\} \\
\{P_3\} & \mapsto \{i : [1, 5], j : [0, \inf]\} \\
\{P_4\} & \mapsto \{i : [1, 5], j : [1, \inf]\} \\
\{P_5\} & \mapsto \{i : [2, 6], j : [1, \inf]\} \\
\{P_6\} & \mapsto \{i : [6, \inf], j : [0, \inf]\}
\end{align*}

\textbf{Example 3:}

\textbf{Input:}  \begin{lstlisting}[basicstyle=\ttfamily]

{P0}
y := 7;
{P1}
while (true) do
    [while_true]
    {P2}
    x := read();
    {P3}
    while (x <= y) do
        [while_true]
        {P4}
        x := x + 1;
        {P5}
    end [while_false] 
    {P6}
end [while_false] 
{P7}
\end{lstlisting}

Initially, the abstract state at $\{P0\}$ is $\{x : [-\inf, \inf], y : [-\inf, \inf]\}$.

\begin{itemize}
\item[1.] Interpret y := 7
    \begin{itemize}
        \item[-] The input abstract state is $\{x : [-\inf, \inf], y : [-\inf, \inf]\}$.
        \item[-] The resulting abstract state is $\{x : [-\inf, \inf], y : [7, 7]\}$.
        \item[-] As a side-effect, the abstract state at $\{P1\}$ is $\{x : [-\inf, \inf], y : [7, 7]\}$.
    \end{itemize}
\item[2.] Interpret the outer while-loop.
    \begin{itemize}
    \item[-] The input abstract state (iteration 0) is $\{x : [-\inf, \inf], y : [7, 7]\}$.
    \item[-] Begin fixed point iteration.
    \item[-] Outer Loop Fixed Point Iteration 1:
        \begin{itemize}
        \item[-] The input abstract state to this iteration is $\{x : [-\inf, \inf], y : [7, 7]\}$.
        \item[-] Filtering the state by true results in the abstract state $\{x : [-\inf, \inf], y : [7, 7]\}$.
          As a side-effect, the abstract state at $\{P2\}$ is $\{x : [-\inf, \inf], y : [7, 7]\}$.
        \item[-] Interpret x := read();
            \begin{itemize}
            \item[-] The input abstract state is $\{x : [-\inf, \inf], y : [7, 7]\}$.
            \item[-] The resulting abstract state is $\{x : [-\inf, \inf], y : [7, 7]\}$.
            \item[-] As a side-effect, the abstract state at $\{P3\}$ is $\{x : [-\inf, \inf], y : [7, 7]\}$.
            \end{itemize}
        \item[-] Interpret the inner while-loop:
            \begin{itemize}
            \item[-] The input abstract state (iteration 0) is $\{x : [-\inf,\inf], y : [7, 7]\}$.
            \item[-] Begin fixed point iteration.
            \item[-] Inner Loop Fixed Point Iteration 1:
                \item[\phantom{-}] \phantom{abt}- The input abstract state to this iteration is $\{x : [-\inf, \inf], y : [7, 7]\}$.
                \item[\phantom{-}] \phantom{abt}- Filtering the state by x <= y results in $\{x : [-\inf, 7], y : [7, 7]\}$.
                    As a side-effect, the abstract state at $\{P4\}$ is $\{x : [-\inf, 7], y : [7, 7]\}$.
                \item[\phantom{-}] \phantom{abt}- Interpret x := x + 1;
                    \item[\phantom{-}] \phantom{abtabt}- The input abstract state is $\{x : [-\inf, 7], y : [7, 7]\}$.
                    \item[\phantom{-}] \phantom{abtabt}- The resulting abstract state is $\{x : [-\inf, 8], y : [7, 7]\}$.
                    \item[\phantom{-}] \phantom{abtabt}- As a side-effect the abstract state at $\{P5\}$ is $\{x : [-\inf, 8], y : [7, 7]\}$
                \item[\phantom{-}] \phantom{abt}- Widen the abstract state from the previous iteration by the interpretation of the loop body
                    \item[\phantom{-}] \phantom{abtabt}- The input abstract state to this iteration is $\{x : [-\inf, \inf], y : [7, 7]\}$.
                    \item[\phantom{-}] \phantom{abtabt}- The result of interpreting the loop body is $\{x : [-\inf, 8], y : [7, 7]\}$.
                    \item[\phantom{-}] \phantom{abtabt}- $\{x : [-\inf, \inf], y : [7, 7]\} \nabla \{x : [-\inf, 8], y : [7, 7]\} = \{x : [-\inf, \inf], y : [7, 7]\}$.
                \item[\phantom{-}] \phantom{abt}- The result of this iteration is $\{x : [-\inf, \inf], y : [7, 7]\}$
            \item[-] We are at a fixed point.  The result of this iteration was the same as the previous one.
            \item[-] Filter the fixed point by the negation of the loop guard, x > y.  Filtering $\{x : [-\inf, \inf], y : [7, 7]\}$ by
            x > y results in $\{x : [8, \inf], y : [7, 7]\}$.  As a side-effect, the abstract state at $\{P6\}$ is $\{x : [8, \inf], y : [7, 7]\}$.
            \end{itemize}
        \item[-] The result of interpreting the inner while loop is $\{x : [8, \inf], y : [7, 7]\}$.
        \item[-] Widen the abstract state from the previous iteration by the interpretation of the loop body
            \begin{itemize}
                \item[-] The input abstract state to this iteration is $\{x : [-\inf, \inf], y : [7, 7]\}$.
                \item[-] The result of interpreting the outer loop body is $\{x : [8, \inf], y : [7, 7]\}$.
                \item[-] $\{x : [-\inf, \inf], y : [7, 7]\} \nabla \{x : [8, \inf], y : [7, 7]\} = \{x : [-\inf, \inf], y : [7, 7]\}$.
            \end{itemize}
        \item[-] The result of this iteration for the outer while loop is $\{x : [-\inf, \inf], y : [7, 7]\}$.
        \end{itemize}
    \item[-] We've reached a fixed point for the outer while loop.  The input state to the first iteration of the fixed point computation for the
      outer loop is the same as the abstract state resulting from the first iteration.
    \item[-] Filter the fixed point for the outer while loop by the negation of the loop guard, false.  Filtering $\{x : [-\inf, \inf], y : [7, 7]\}$
      by false results in $\{x : \bot, y : \bot\}$.  As a side-effect, the abstract state at $\{P7\}$ is set to $\{x : \bot, y : \bot\}$.
    \end{itemize}
\end{itemize}

There are no more statements to interpret, and the answer is

\begin{align*}
\{P_0\} & \mapsto \{x : [-\inf, \inf], y : [-\inf, \inf]\} \\
\{P_1\} & \mapsto \{x : [-\inf, \inf], y : [7, 7]\} \\
\{P_2\} & \mapsto \{x : [-\inf, \inf], y : [7, 7]\} \\
\{P_3\} & \mapsto \{x : [-\inf, \inf], y : [7, 7]\} \\
\{P_4\} & \mapsto \{x : [-\inf, 7], y : [7, 7]\} \\
\{P_5\} & \mapsto \{x : [-\inf, 8], y : [7, 7]\} \\
\{P_6\} & \mapsto \{x : [8, \inf], y : [7, 7]\} \\
\{P_7\} & \mapsto \{x : \bot, y : \bot\}
\end{align*}

Now, please solve this, outputting the intermediary steps you take:

\textbf{[Input Program]}

\end{tcolorbox}

\subsection{Transitional Strategy}
\label{appendix:transitional}
\begin{tcolorbox}[breakable, colback=gray!20, colframe=black]
\textbf{Context:} 

Given a program, analyze the program with abstract interpretation, using the interval abstract domain.
Programs are composed of assignment, skip, if-then-else, while-loops, and sequential composition of these statements,
where program variables are integer variables. The goal is to output an abstract state for each program location.  
An abstract state maps each program variable to an interval, or the empty interval $\bot$. For example, 
$\{x : [1, 4], y : [-1, 3]\}$ means that x can take on values between 1 and 4 and y can take on values between -1 and 3.
$\bot$ means that the variable cannot have any concrete value. \\

Each abstract state should be sound.  For instance if the abstract state at location $\{P\}$ maps x to $[4, 10]$, then in
any concrete execution of the program, the value of x should be between 4 and 10 at location $\{P\}$. \\ 

Arithmetic expressions are interpreted with interval arithmetic.  Be cautious of edge cases in interpreting division
with interval arithmetic.  For example, $[1, 3] / [0, 0] = \bot$, as no valid value results from a division by 0. 
Furthermore, $[1, 3] / [-2, 3] = [-inf, inf]$, as division by 0 may or may not occur. \\

\texttt{read()} expressions are interpreted as $[-inf, inf]$, as reading from the standard input can result in any value. \\

You should abstractly interpret programs by first deriving a set of fixed point equations, where each program location
corresponds to one equation. Then, solve the fixed point equations iteratively until you reach a fixed point.
The fixed point equation associated with the location at program entry, $\{P0\}$, maps each program variable to 
$[-\inf, \inf]$, indicating that at the beginning of the program, the variables can have any integer value. \\

There are several directives in the annotated programs that help keep track of control flow, as well as indicate
how the fixed point equations should be defined. \\

$\mathtt{[if\_then]}$ means that the fixed point equation corresponding to the location after the directive is the result of filtering
the abstract state at the location corresponding to the input of the if-then-else statement, by the if guard. \\
$\mathtt{[if\_else]}$ means that the fixed point equation corresponding to the location after the directive is the result of filtering
the abstract state at the location corresponding to the input of the if-then-else statement, by the negation of the if guard. \\
$\mathtt{[if\_end]}$ means the fixed point equation corresponding to the location after the directive is the result of joining 
the abstract states at the locations of the end of each branch in the if-statement. \\

$\mathtt{[while\_true]}$ means that the fixed point equation at the location after the directive first joins the 
abstract states at the program locations before the while-loop and after the last statement
in the loop body, and filters this result by the loop guard. \\

$\mathtt{[while\_false]}$ means that the fixed point equation at the location after the directive first joins the 
abstract states at the program locations before the while-loop and after the last statement
in the loop body, and filters this result by the negation of the loop guard. \\ 

Some examples of filtering are:
\begin{itemize}
    \item[-] Filtering abstract state ${x : [5, 7], y : [6, 8]}$ by $\mathtt{!(read() == 0)}$ results in the same abstract state, because
    we cannot know for certain if the result of reading from standard input is $0$.
    \item[-] Filtering abstract state $\{x : [5, 10], y : [5, inf]\}$ by $\mathtt{!(y == 6)}$ results in ${x : [5, 10], y : [5, inf]}$.  Filtering
    by $\mathtt{!(y == 6)}$ is equivalent to filtering by $y > 6 || y < 6$.  Filtering the abstract state by $y > 6$ results in 
    ${x : [5, 10], y : [7, \inf]}$.  Filtering the abstract state by $y < 6$ results in ${x : [5, 10], y : [5, 5]}$. 
    Joining the resulting abstract states results in ${x : [5, 10], y : [5, inf]}$.
    \item[-] Filtering abstract state ${x : [5, 9], y : [10, 12]}$ by $\mathtt{y == 16}$ results in $\{x : \bot, y : \bot\}$, as it is impossible for
    y to be $16$.
    \item[-] Filtering the abstract state ${x : [5, 10], y : [4, 9]}$ by $\mathtt{(y <= 8) \&\& (x <= y)}$ results in the 
    filtering the state by $\mathtt{y <= 8}$ and filtering the state by $x <= y$ and then intersecting the resulting states.
    Filtering $\{x : [5, 10], y : [4, 9]\}$ by $y <= 8 $results in$ {x : [5, 10], y : [4, 8]}.$  Filtering
   $ {x : [5, 10], y : [4, 9]}$ by $x <= y $ results in ${x : [5, 9], y : [4, 9]}$.  Intersecting both states 
    results in ${x : [5, 9], y : [4, 8]}$.
\end{itemize}

In the equations, use Interpret(assignment, S) and Interpret(skip, S) to denote interpreting the result of applying
an assignment statement to abstract state S and applying a skip statement to abstract state S, respectively.
Use Filter(B, S) to filter abstract state S by boolean expression B.

Once the equations are set up, fixed point computation is conducted using a worklist algorithm.  Initially, all program
locations are added to the worklist.  If a location is in the worklist, this indicates that the abstract state at that
location has not stabilized yet.  When we compute the abstract state at a location just after a $\mathtt{[while\_true]}$
directive, we widen the result of computation by the previous abstract state at the same location.  This ensures
termination of the analysis.  The widening of two intervals is defined symbolically as $[a, b] \nabla [c, d] = [\text{if  } c < a, \text{ then } -\inf \text{ else } a, \text{ if } d > b \text{ then } \inf \text{ else } b]$.  
For example, $[6, 7] \nabla [9, 10] = [6, \inf]$. Note that $\bot \nabla [c, d] = [c, d]$ and $[a, b] \nabla \bot = [a, b]$.

A solution for the fixed point equations is reached after the worklist is empty, and the final abstract states are returned.

Here are some examples:

\textbf{Example 1:}

\begin{lstlisting}[basicstyle=\ttfamily]
Input: {P0} 
    x := read();
    {P1}
    if (x < 3) then 
        [if_then] 
        {P2} 
        x := x - 1; 
        {P3} 
        x := x * 2; 
        {P4} 
    else 
        [if_else] 
        {P5} 
        x := x + 2; 
        {P6}
    end [endif] {P7}
\end{lstlisting}

\textbf{Output:}

1. Create a system of fixed point equations. \\

Here, $M(\{P\})$ denotes the abstract state at location $\{P\}$.  $M(\{P\})(x)$ denotes the abstract value
of variable x at location $\{P\}$. $F_0$ denotes the fixed point equation at $\{P0\}$.

\begin{align*}
F_0(M) &= \{x : [-\inf, \inf]\} \\
F_1(M) &= \mathtt{Interpret}(\mathtt{x := read()}, M(\{P_0\})) \\
F_2(M) &= \mathtt{Filter}(\mathtt{x < 3}, M(\{P_1\})) \\
F_3(M) &= \mathtt{Interpret}(\mathtt{x := x - 1}, M(\{P_2\})) \\
F_4(M) &= \mathtt{Interpret}(\mathtt{x := x * 2}, M(\{P_3\})) \\
F_5(M) &= \mathtt{Filter}(\mathtt{x >= 3}, M(\{P_1\})) \\
F_6(M) &= \mathtt{Interpret}(\mathtt{x := x + 2}, M(\{P_5\})) \\
F_7(M) &= M(\{P_4\}) \sqcup M(\{P_6\})
\end{align*}

2.  Solve the fixed point equations using a worklist algorithm.

Initially, the map of program locations to abstract states looks like: 
\begin{align*}
M(\{P0\}) &= \{x : \bot\}, \\
M(\{P1\}) &= \{x : \bot\}, \\
M(\{P2\}) &= \{x : \bot\}, \\
M(\{P3\}) &= \{x : \bot\}, \\
M(\{P4\}) &= \{x : \bot\}, \\
M(\{P5\}) &= \{x : \bot\}, \\
M(\{P6\}) &= \{x : \bot\}, \\
M(\{P7\}) &= \{x : \bot\}.
\end{align*}

The worklist W is $\{\{P0\}, \{P1\}, \{P2\}, \{P3\}, \{P4\}, \{P5\}, \{P6\}, \{P7\}$.

\begin{itemize}
\item Pick $\{P_0\}$ from $W$.
   \begin{itemize}
   \item Remove $\{P_0\}$ from $W$.
   \item $M(\{P_0\})$ is $\{x : \bot\}$.
   \item Compute $F_0(M)$, and update the value of $M(\{P_0\})$, resulting in $M(\{P_0\}) = \{x : [-\inf,  \inf]\}$.
   \item $M(\{P_0\})$ has changed, so add the program locations whose fixed point equations directly depend on $M(\{P_0\})$ to $W$.
       \begin{itemize}
       \item Add $\{P_1\}$ to $W$.
       \end{itemize}
   \item $W$ is now $\{\{P_1\}, \{P_2\}, \{P_3\}, \{P_4\}, \{P_5\}, \{P_6\}, \{P_7\}\}$.
   \end{itemize}
\end{itemize}

\begin{itemize}
\item Pick $\{P_1\}$ from $W$.
    \begin{itemize}
    \item Remove $\{P_1\}$ from $W$.
    \item $M(\{P_1\})$ is $\{x : \bot\}$.
    \item Compute $F_1(M)$, and update the value of $M(\{P_1\})$, resulting in $M(\{P_1\}) = \{x : [-\inf,  \inf]\}$, where
        \begin{itemize}
        \item $M(\{P_1\})(x) = [-\inf, \inf]$ is the result of interpreting $\mathtt{x := read()}$.
        \end{itemize}
    \item $M(\{P_1\})$ has changed, so add the program locations whose fixed point equations directly depend on $M(\{P_1\})$ to $W$.
        \begin{itemize}
        \item Add $\{P_2\}$ and $\{P_5\}$ to $W$.
        \end{itemize}
    \item $W$ is now $\{\{P_2\}, \{P_3\}, \{P_4\}, \{P_5\}, \{P_6\}, \{P_7\}\}$.
    \end{itemize}
\end{itemize}

\begin{itemize}
\item Pick $\{P_2\}$ from $W$.
   \begin{itemize}
   \item Remove $\{P_2\}$ from $W$.
   \item $M(\{P_2\})$ is $\{x : \bot\}$.
   \item Compute $F_2(M)$:
       \begin{itemize}
       \item $M(\{P_1\}) = \{x : [-\inf, \inf]\}$
       \item Filtering $M(\{P_1\})$ by $x < 3$ results in:
           \begin{itemize}
           \item $\{x : [-\inf, 2]\}$
           \end{itemize}
       \item Update $M(\{P_2\})$ to be $\{x : [-\inf, 2]\}$.
       \end{itemize}
   \item $M(\{P_2\})$ has changed, so add the program locations whose fixed point equations directly depend on $M(\{P_2\})$ to $W$.
       \begin{itemize}
       \item Add $\{P_3\}$ to $W$.
       \end{itemize}
   \item $W$ is now $\{\{P_3\}, \{P_4\}, \{P_5\}, \{P_6\}, \{P_7\}\}$.
   \end{itemize}
\end{itemize}

\begin{itemize}
\item[-] Pick $\{P_3\}$ from $W$.
   \begin{itemize}
   \item[-] Remove $\{P_3\}$ from $W$.
   \item[-] $M(\{P_3\})$ is $\{x : \bot\}$.
   \item[-] Compute $F_3(M)$ and update the value of $M(\{P_3\})$, which results in $M(\{P_3\}) = \{x : [-\inf, 1]\}$, where 
       \begin{itemize}
       \item[-] $M(\{P_3\})(x) = M(\{P_2\})(x) - [1, 1]$
                    $= [-\inf, 2] - [1, 1]$
                   $= [-\inf, 1]$
       \end{itemize}
   \item[-] $M(\{P_3\})$ has changed, so add the program locations whose fixed point equations directly depend on $M(\{P_3\})$ to $W$.
       \begin{itemize}
       \item[-] Add $\{P_4\}$ to $W$.
       \end{itemize}
   \item[-] $W$ is now $\{\{P_4\}, \{P_5\}, \{P_6\}, \{P_7\}\}$.
   \end{itemize}
\end{itemize}

\begin{itemize}
\item[-] Pick $\{P4\}$ from $W$.
    \begin{itemize}
    \item[-] Remove $\{P4\}$ from $W$.
    \item[-] $M(\{P4\})$ is $\{x : \bot\}$.
    \item[-] Compute $F_4(M)$ and update the value of $M(\{P4\})$, which results in $M(\{P4\}) = \{x : [-\inf, 2]\}$, where 
        \begin{itemize}
        \item[-] $M(\{P4\})(x) = M(\{P3\})(x) * [2, 2]$
                    $= [-\inf, 1] * [2, 2]$
                    $= [-\inf, 2]$
        \end{itemize}
    \item[-] $M(\{P4\})$ has changed, so add the program locations whose fixed point equations directly depend on $M(\{P4\})$ to $W$.
        \begin{itemize}
        \item[-] Add $\{P7\}$ to $W$.
        \end{itemize}
    \item[-] $W$ is now $\{\{P5\}, \{P6\}, \{P7\}\}$.
    \end{itemize}
\end{itemize}

\begin{itemize}
\item[-] Pick $\{P5\}$ from $W$.
   \begin{itemize}
   \item[-] Remove $\{P5\}$ from $W$.
   \item[-] $M(\{P5\})$ is $\{x : \bot\}$.
   \item[-] Compute $F_5(M)$: 
       \begin{itemize}
       \item[-] $M(\{P1\}) = \{x : [-\inf, \inf]\}$
       \item[-] Filtering $M(\{P1\})$ by $x \geq 3$ results in:
           \begin{itemize}
           \item[-] $\{x : [3, \inf]\}$
           \end{itemize}
       \item[-] Update $M(\{P5\})$ to be $\{x : [3, \inf]\}$
       \end{itemize}
   \item[-] $M(\{P5\})$ has changed, so add the program locations whose fixed point equations directly depend on $M(\{P5\})$ to $W$.
       \begin{itemize}
       \item[-] Add $\{P6\}$ to $W$.
       \end{itemize}
   \item[-] $W$ is now $\{\{P6\}, \{P7\}\}$.
   \end{itemize}
\end{itemize}

\begin{itemize}
\item[-] Pick $\{P6\}$ from $W$.
   \begin{itemize}
   \item[-] Remove $\{P6\}$ from $W$.
   \item[-] $M(\{P6\})$ is $\{x : \bot\}$.
   \item[-] Compute $F_6(M)$ and update the value of $M(\{P6\})$, which results in $M(\{P6\}) = \{x : [5, \inf]\}$, where 
       \begin{itemize}
       \item[-] $M(\{P6\})(x) = M(\{P5\})(x) + [2, 2]$
                   $= [3, \inf] + [2, 2]$
                   $= [5, \inf]$
       \end{itemize}
   \item[-] $M(\{P6\})$ has changed, so add the program locations whose fixed point equations directly depend on $M(\{P6\})$ to $W$.
       \begin{itemize}
       \item[-] Add $\{P7\}$ to $W$.
       \end{itemize}
   \item[-] $W$ is now $\{\{P7\}\}$.
   \end{itemize}
\end{itemize}

\begin{itemize}
\item[-] Pick $\{P7\}$ from $W$.
   \begin{itemize}
   \item[-] Remove $\{P7\}$ from $W$.
   \item[-] $M(\{P7\})$ is $\{x : \bot\}$.
   \item[-] Compute $F_7(M)$:
       \begin{itemize}
       \item[-] $M(\{P4\}) \sqcup M(\{P6\}) = \{x : [-\inf, 2]\} \sqcup \{x : [5, \inf]\}$
                           $= \{x : [-\inf, \inf]\}$
       \item[-] Update $M(\{P7\})$ to be $\{x : [-\inf, \inf]\}$
       \end{itemize}
   \item[-] $M(\{P7\})$ has changed so add the program locations whose fixed point equations directly depend on $M(\{P7\})$ to $W$.
       \begin{itemize}
       \item[-] According to the system of equations, there is no such location, so no location is added to $W$.
       \end{itemize}
   \item[-] $W$ is now $\{\}$.
   \end{itemize}
\end{itemize}

The worklist is empty, meaning we've finished the analysis and M is

\[
M(\{P0\}) = \{x : [-\inf, \inf]\}
\]
\[
M(\{P1\}) = \{x : [-\inf, \inf]\}
\]
\[
M(\{P2\}) = \{x : [-\inf, 2]\}
\]
\[
M(\{P3\}) = \{x : [-\inf, 1]\}
\]
\[
M(\{P4\}) = \{x : [-\inf, 2]\}
\]
\[
M(\{P5\}) = \{x : [3, \inf]\}
\]
\[
M(\{P6\}) = \{x : [5, \inf]\}
\]
\[
M(\{P7\}) = \{x : [-\inf, \inf]\}
\]

\textbf{Example 2:}

\textbf{Input:}

\begin{lstlisting}[basicstyle=\ttfamily]
{P0}
i := 1;
{P1}
j := 0;
{P2} 
while (i <= 5) do 
    [while_true]
    {P3} 
    j := j + i;
    {P4} 
    i := i + 1; 
    {P5} 
end [while_false]
{P6}
\end{lstlisting}

\textbf{Output:}

1. Create a system of fixed point equations.

Here, $M(\{P\})$ denotes the abstract state at location $\{P\}$.  $M(\{P\})(x)$ denotes the abstract value
of variable x at location $\{P\}$.

\[
F_0(M) = \{i : [-\inf, \inf], j : [-\inf, \inf]\}
\]
\[
F_1(M) = \text{Interpret}(i := 1, M(\{P_0\}))
\]
\[
F_2(M) = \text{Interpret}(j := 0, M(\{P_1\}))
\]
\[
F_3(M) = \text{Filter}(i \leq 5, M(\{P_2\}) \sqcup M(\{P_5\}))
\]
\[
F_4(M) = \text{Interpret}(j := j + i, M(\{P_3\}))
\]
\[
F_5(M) = \text{Interpret}(i := i + 1, M(\{P_4\}))
\]
\[
F_6(M) = \text{Filter}(i > 5, M(\{P_2\}) \sqcup M(\{P_5\}))
\]

2. Solve the fixed point equations using a worklist algorithm.

Initially, the map of program locations to abstract states looks like: 
\[
M(\{P_0\}) = \{i : \bot, j : \bot\}
\]
\[
M(\{P_1\}) = \{i : \bot, j : \bot\}
\]
\[
M(\{P_2\}) = \{i : \bot, j : \bot\}
\]
\[
M(\{P_3\}) = \{i : \bot, j : \bot\}
\]
\[
M(\{P_4\}) = \{i : \bot, j : \bot\}
\]
\[
M(\{P_5\}) = \{i : \bot, j : \bot\}
\]
\[
M(\{P_6\}) = \{i : \bot, j : \bot\}
\]

The worklist W is $\{\{P0\}, \{P1\}, \{P2\}, \{P3\}, \{P4\}, \{P5\}, \{P6\}\}$.

\begin{itemize}
\item[-] Pick $\{P0\}$ from $W$.
   \begin{itemize}
   \item[-] Remove $\{P0\}$ from $W$.
   \item[-] $M(\{P0\})$ is $\{i : \bot, j : \bot\}$.
   \item[-] Compute $F_0(M)$, and update the value of $M(\{P0\})$, resulting in $M(\{P0\}) = \{i : [-\inf, \inf], j : [-\inf, \inf]\}$.
   \item[-] $M(\{P0\})$ has changed, so add the program locations whose fixed point equations directly depend on $M(\{P0\})$ to $W$.
       \begin{itemize}
       \item[-] Add $\{P1\}$ to $W$.
       \end{itemize}
   \item[-] $W$ is now $\{\{P1\}, \{P2\}, \{P3\}, \{P4\}, \{P5\}, \{P6\}\}$.
   \end{itemize}
\end{itemize}

\begin{itemize}
\item[-] Pick $\{P1\}$ from $W$.
   \begin{itemize}
   \item[-] Remove $\{P1\}$ from $W$.
   \item[-] $M(\{P1\})$ is $\{i : \bot, j : \bot\}$.
   \item[-] Compute $F_1(M)$, and update the value of $M(\{P1\})$, resulting in $M(\{P1\}) = \{i : [1, 1], j : [-\inf, \inf]\}$, where
       \begin{itemize}
       \item[-] $M(\{P1\})(i) = [1, 1]$
       \item[-] $M(\{P1\})(j) = M(\{P0\})(j)$
       \end{itemize}
   \item[-] $M(\{P1\})$ has changed, so add the program locations whose fixed point equations directly depend on $M(\{P1\})$ to $W$.
       \begin{itemize}
       \item[-] Add $\{P2\}$ to $W$.
       \end{itemize}
   \item[-] $W$ is now $\{\{P2\}, \{P3\}, \{P4\}, \{P5\}, \{P6\}\}$.
   \end{itemize}
\end{itemize}

\begin{itemize}
\item[-] Pick $\{P2\}$ from $W$.
   \begin{itemize}
   \item[-] Remove $\{P2\}$ from $W$.
   \item[-] $M(\{P2\})$ is $\{i : \bot, j : \bot\}$.
   \item[-] Compute $F_2(M)$ and update the value of $M(\{P2\})$, resulting in $M(\{P2\}) = \{i : [1 , 1], j : [0, 0]\}$, where
       \begin{itemize}
       \item[-] $M(\{P2\})(i) = M(\{P1\})(i)$
       \item[-] $M(\{P2\})(j) = [0, 0]$
       \end{itemize}
   \item[-] $M(\{P2\})$ has changed, so add the program locations whose fixed point equations directly depend on $M(\{P2\})$ to $W$.
       \begin{itemize}
       \item[-] Add $\{P3\}$ and $\{P6\}$ to $W$.
       \end{itemize}
   \item[-] $W$ is now $\{\{P3\}, \{P4\}, \{P5\}, \{P6\}\}$.
   \end{itemize}
\end{itemize}

\begin{itemize}
\item[-] Pick $\{P3\}$ from $W$.
   \begin{itemize}
   \item[-] Remove $\{P3\}$ from $W$.
   \item[-] $M(\{P3\})$ is $\{i : \bot, j : \bot\}$.
   \item[-] Compute $F_3(M)$: 
       \begin{itemize}
       \item[-] $M(\{P2\}) \sqcup M(\{P5\}) = \{i : [1 , 1], j : [0, 0]\} \sqcup \{i : \bot, j : \bot\} = \{i : [1 , 1], j : [0, 0]\}$.
       \item[-] Filtering $\{i : [1 , 1], j : [0, 0]\}$ by $i \leq 5$ results in:
           \begin{itemize}
           \item[-] $S = \{i : [1 , 1], j : [0, 0]\}$
           \end{itemize}
       \end{itemize}
   \item[-] Because $\{P3\}$ corresponds to a loop head, we widen $M(\{P3\})$ by $S$.
       \begin{itemize}
       \item[-] $M(\{P3\}) \nabla S$ results in $S' = \{i : [1, 1], j : [0, 0]\}$, where
           \begin{itemize}
           \item[-] $S'(i) = \bot \nabla [1, 1]$
           \item[-] $S'(j) = \bot \nabla [0, 0]$
           \end{itemize}
       \end{itemize}
   \item[-] Update $M(\{P3\})$ to $\{i : [1, 1], j : [0, 0]\}$.
   \item[-] $M(\{P3\})$ has changed, so add the program locations whose fixed point equations directly depend on $M(\{P3\})$ to $W$.
       \begin{itemize}
       \item[-] Add $\{P4\}$ to $W$.
       \end{itemize}
   \item[-] $W$ is now $\{\{P4\}, \{P5\}, \{P6\}\}$
   \end{itemize}
\end{itemize}

\begin{itemize}
\item[-] Pick $\{P4\}$ from $W$.
   \begin{itemize}
   \item[-] Remove $\{P4\}$ from $W$.
   \item[-] $M(\{P4\})$ is $\{i : \bot, j : \bot\}$.
   \item[-] Compute $F_4(M)$, and update the value of $M(\{P4\})$, resulting in $M(\{P4\}) = \{i : [1, 1], j : [1, 1]\}$, where
       \begin{itemize}
       \item[-] $M(\{P4\})(i) = M(\{P3\})(i)$
                    $= [1, 1]$
       \item[-] $M(\{P4\})(j) = M(\{P3\})(j) + M(\{P3\})(i)$
                   $= [0, 0] + [1, 1]$
                   $= [1, 1]$
       \end{itemize}
   \item[-] $M(\{P4\})$ has changed, so add the program locations whose fixed point equations directly depend on $M(\{P4\})$ to $W$.
       \begin{itemize}
       \item[-] Add $\{P5\}$ to $W$.
       \end{itemize}
   \item[-] $W$ is now $\{\{P5\}, \{P6\}\}$.
   \end{itemize}
\end{itemize}

\begin{itemize}
\item[-] Pick $\{P5\}$ from $W$.
   \begin{itemize}
   \item[-] Remove $\{P5\}$ from $W$.
   \item[-] $M(\{P5\})$ is $\{i : \bot, j : \bot\}$.
   \item[-] Compute $F_5(M)$, and update the value of $M(\{P5\})$, resulting in $M(\{P5\}) = \{i : [2, 2], j : [1, 1]\}$, where
       \begin{itemize}
       \item[-] $M(\{P5\})(i) = M(\{P4\})(i) + [1, 1]$
                   $= [1, 1] + [1, 1]$
                   $= [2, 2]$
       \item[-] $M(\{P5\})(j) = M(\{P4\})(j)$
       \end{itemize}
   \item[-] $M(\{P5\})$ has changed, so add the program locations whose fixed point equations directly depend on $M(\{P5\})$ to $W$.
       \begin{itemize}
       \item[-] Add $\{P3\}$ and $\{P6\}$ to $W$.
       \end{itemize}
   \item[-] $W$ is now $\{\{P3\}, \{P6\}\}$.
   \end{itemize}
\end{itemize}

\begin{itemize}
\item[-] Pick $\{P3\}$ from $W$.
   \begin{itemize}
   \item[-] Remove $\{P3\}$ from $W$.
   \item[-] $M(\{P3\})$ is $\{i : [1, 1], j : [0, 0]\}$.
   \item[-] Compute $F_3(M)$:
       \begin{itemize}
       \item[-] $M(\{P2\}) \sqcup M(\{P5\}) = \{i : [1 , 1], j : [0, 0]\} \sqcup \{i : [2, 2], j : [1, 1]\} = \{i : [1 , 2], j : [0 , 1]\}$
       \item[-] Filtering $\{i : [1 , 2], j : [0 , 1]\}$ by $i \leq 5$ results in:
           \begin{itemize}
           \item[-] $S = \{i : [1 , 2], j : [0 , 1]\}$
           \end{itemize}
       \end{itemize}
   \item[-] Because $\{P3\}$ corresponds to a loop head, we widen $M(\{P3\})$ by $S$.
       \begin{itemize}
       \item[-] $M(\{P3\}) \nabla S$ results in $S' = \{i : [1, \inf], j : [0, \inf]\}$, where
           \begin{itemize}
           \item[-] $S'(i) = [1, 1] \nabla [1, 2]$
                   $= [1, \inf]$
           \item[-] $S'(j) = [0, 0] \nabla [0, 1]$
                   $= [0, \inf]$
           \end{itemize}
       \end{itemize}
   \item[-] Update $M(\{P3\})$ to $\{i : [1, \inf], j : [0, \inf]\}$.
   \item[-] $M(\{P3\})$ has changed, so add the program locations whose fixed point equations directly depend on $M(\{P3\})$ to $W$.
       \begin{itemize}
       \item[-] Add $\{P4\}$ to $W$.
       \end{itemize}
   \item[-] $W$ is now $\{\{P4\}, \{P6\}\}$
   \end{itemize}
\end{itemize}

\begin{itemize}
\item[-] Pick $\{P4\}$ from $W$.
   \begin{itemize}
   \item[-] Remove $\{P4\}$ from $W$.
   \item[-] $M(\{P4\})$ is $\{i : [1, 1], j : [1, 1]\}$.
   \item[-] Compute $F_4(M)$ and update the value of $M(\{P4\})$, resulting in $M(\{P4\}) = \{i : [1, \inf], j : [1, \inf]\}$, where
       \begin{itemize}
       \item[-] $M(\{P4\})(i) = M(\{P3\})(i)$
                   $= [1, \inf]$
       \item[-] $M(\{P4\})(j) = M(\{P3\})(j) + M(\{P3\})(i)$
                   $= [0, \inf] + [1, \inf]$
                   $= [1, \inf]$
       \end{itemize}
   \item[-] $M(\{P4\})$ has changed, so add the program locations whose fixed point equations directly depend on $M(\{P4\})$ to $W$.
       \begin{itemize}
       \item[-] Add $\{P5\}$ to $W$.
       \end{itemize}
   \item[-] $W$ is now $\{\{P5\}, \{P6\}\}$.
   \end{itemize}
\end{itemize}

\begin{itemize}
\item[-] Pick $\{P5\}$ from $W$.
    \begin{itemize}
    \item[-] Remove $\{P5\}$ from $W$.
    \item[-] $M(\{P5\})$ is $\{i : [2, 2], j : [1, 1]\}$.
    \item[-] Compute $F_5(M)$, and update the value of $M(\{P5\})$, resulting in $M(\{P5\}) = \{i : [2, \inf], j : [1, \inf]\}$, where
        \begin{itemize}
        \item[-] $M(\{P5\})(i) = M(\{P4\})(i) + [1, 1]$
                                 $= [1, \inf] + [1, 1]$
                                 $= [2, \inf]$
        \item[-] $M(\{P5\})(j) = M(\{P4\})(j)$
                                 $= [1, \inf]$
        \end{itemize}
    \item[-] $M(\{P5\})$ has changed, so add the program locations whose fixed point equations directly depend on $M(\{P5\})$ to $W$.
        \begin{itemize}
        \item[-] Add $\{P3\}$ and $\{P6\}$ to $W$.
        \end{itemize}
    \item[-] $W$ is now $\{\{P3\}, \{P6\}\}$.
    \end{itemize}
\end{itemize}

\begin{itemize}
\item[-] Pick $\{P3\}$ from $W$.
    \begin{itemize}
    \item[-] Remove $\{P3\}$ from $W$.
    \item[-] $M(\{P3\})$ is $\{i : [1, \inf], j : [0, \inf]\}$.
    \item[-] Compute $F_3(M)$:
        \begin{itemize}
        \item[-] $M(\{P2\}) \sqcup M(\{P5\}) = \{i : [1, 1], j : [0, 0]\} \sqcup \{i : [2, \inf], j : [1, \inf]\} = \{i : [1, \inf], j : [0, \inf]\}$
        \item[-] Filtering $\{i : [1, \inf], j : [0, \inf]\}$ by $i \leq 5$ results in:
            \begin{itemize}
            \item[-] $S = \{i : [1, 5], j : [0, \inf]\}$
            \end{itemize}
        \end{itemize}
    \item[-] Because $\{P3\}$ corresponds to a loop head, we widen $M(\{P3\})$ by $S$.
        \begin{itemize}
        \item[-] $M(\{P3\}) \nabla S$ results in $S' = \{i : [1, \inf], j : [0, \inf]\}$, where
            \begin{itemize}
            \item[-] $S'(i) = [1, \inf] \nabla [1, 5]$
                            $= [1, \inf]$
            \item[-] $S'(j) = [0, \inf] \nabla [0, \inf]$
                            $= [0, \inf]$
            \end{itemize}
        \item[-] Now, $M(\{P3\}) = \{i : [1, \inf], j : [0, \inf]\}$.
        \end{itemize}
    \item[-] $M(\{P3\})$ has not changed, so do not add anything to the worklist.
    \item[-] $W$ is now $\{\{P6\}\}$.
    \end{itemize}
\end{itemize}

\begin{itemize}
\item[-] Pick $\{P6\}$ from $W$.
    \begin{itemize}
    \item[-] Remove $\{P6\}$ from $W$.
    \item[-] $M(\{P6\})$ is $\{i : \bot, j : \bot\}$.
    \item[-] Compute $F_6(M)$:
        \begin{itemize}
        \item[-] $M(\{P2\}) \sqcup M(\{P5\}) = \{i : [1, 1], j : [0, 0]\} \sqcup \{i : [2, \inf], j : [1, \inf]\} = \{i : [1, \inf], j : [0, \inf]\}$
        \item[-] Filtering $\{i : [1, \inf], j : [0, \inf]\}$ by $i > 5$ results in
            \begin{itemize}
            \item[-] $\{i : [6, \inf], j : [0, \inf]\}$
            \end{itemize}
        \item[-] Now, $M(\{P6\}) = \{i : [6, \inf], j : [0, \inf]\}$
        \end{itemize}
    \item[-] $M(\{P6\})$ has changed, so add the program locations whose fixed point equations directly depend on $M(\{P6\})$ to $W$.
        \begin{itemize}
        \item[-] According to the system of equations, there is no such location, so no location is added to $W$.
        \end{itemize}
    \item[-] $W$ is now $\{\}$.
    \end{itemize}
\end{itemize}

The worklist is empty, meaning we've finished the analysis and M is 

\[
M(\{P_0\}) = \{i : [-\inf, \inf], j : [-\inf, \inf]\}
\]
\[
M(\{P_1\}) = \{i : [1, 1], j : [-\inf, \inf]\}
\]
\[
M(\{P_2\}) = \{i : [1, 1], j : [0, 0]\}
\]
\[
M(\{P_3\}) = \{i : [1, \inf], j : [0, \inf]\}
\]
\[
M(\{P_4\}) = \{i : [1, \inf], j : [1, \inf]\}
\]
\[
M(\{P_5\}) = \{i : [2, \inf], j : [1, \inf]\}
\]
\[
M(\{P_6\}) = \{i : [6, \inf], j : [0, \inf]\}
\]

\textbf{Example 3:}

\textbf{Input:}

\begin{lstlisting}[basicstyle=\ttfamily]
{P0}
y := 7;
{P1}
while (true) do
    [while_true]
    {P2}
    x := read();
    {P3}
    while (x <= y) do
        [while_true]
        {P4}
        x := x + 1;
        {P5}
    end [while_false] 
    {P6}
end [while_false] 
{P7}
\end{lstlisting}

1. Create a system of fixed point equations.

Here, $M(\{P\})$ denotes the abstract state at location $\{P\}$.  $M(\{P\})(x)$ denotes the abstract value
of variable x at location $\{P\}$.

\[
F_0(M) = \{x : [-\inf, \inf], y : [-\inf, \inf]\}
\]
\[
F_1(M) = \text{Interpret}(y := 7, M(\{P_0\}))
\]
\[
F_2(M) = \text{Filter}(\text{true}, M(\{P_1\}) \sqcup M(\{P_6\}))
\]
\[
F_3(M) = \text{Interpret}(x := \text{read}(), M(\{P_2\}))
\]
\[
F_4(M) = \text{Filter}(x \leq y, M(\{P_3\}) \sqcup M(\{P_5\}))
\]
\[
F_5(M) = \text{Interpret}(x := x + 1, M(\{P_4\}))
\]
\[
F_6(M) = \text{Filter}(x > y, M(\{P_3\}) \sqcup M(\{P_5\}))
\]
\[
F_7(M) = \text{Filter}(\text{false}, M(\{P_1\}) \sqcup M(\{P_6\}))
\]

2.  Solve the fixed point equations using a worklist algorithm.

Initially, the map of program locations to abstract states looks like: 
\[
M(\{P0\}) = \{x : \bot, y : \bot\}
\]
\[
M(\{P1\}) = \{x : \bot, y : \bot\}
\]
\[
M(\{P2\}) = \{x : \bot, y : \bot\}
\]
\[
M(\{P3\}) = \{x : \bot, y : \bot\}
\]
\[
M(\{P4\}) = \{x : \bot, y : \bot\}
\]
\[
M(\{P5\}) = \{x : \bot, y : \bot\}
\]
\[
M(\{P6\}) = \{x : \bot, y : \bot\}
\]
\[
M(\{P7\}) = \{x : \bot, y : \bot\}
\]

The worklist W is $\{P0\}, \{P1\}, \{P2\}, \{P3\}, \{P4\}, \{P5\}, \{P6\}, \{P7\}\}$.

\begin{itemize}
\item[-] Pick $\{P0\}$ from $W$.
    \begin{itemize}
    \item[-] Remove $\{P0\}$ from $W$.
    \item[-] $M(\{P0\})$ is $\{i : \bot, j : \bot\}$.
    \item[-] Compute $F_0(M)$, and update the value of $M(\{P0\})$, resulting in $M(\{P0\}) = \{x : [-\inf, \inf], y : [-\inf, \inf]\}$
    \item[-] $M(\{P0\})$ has changed, so add the program locations whose fixed point equations directly depend on $M(\{P0\})$ to $W$.
        \begin{itemize}
        \item[-] Add $\{P1\}$ to $W$.
        \end{itemize}
    \item[-] $W$ is now $\{\{P1\}, \{P2\}, \{P3\}, \{P4\}, \{P5\}, \{P6\}, \{P7\}\}$.
    \end{itemize}
\end{itemize}

\begin{itemize}
\item[-] Pick $\{P1\}$ from $W$.
    \begin{itemize}
    \item[-] Remove $\{P1\}$ from $W$.
    \item[-] $M(\{P1\})$ is $\{x : \bot, y : \bot\}$.
    \item[-] Compute $F_1(M)$ and update the value of $M(\{P1\})$, resulting in $M(\{P1\}) = \{x : [-\inf, \inf], y : [7, 7]\}$, where
        \begin{itemize}
        \item[-] $M(\{P1\})(x) = M(\{P0\})(x)$
        \item[-] $M(\{P1\})(y) = [7, 7]$
        \end{itemize}
    \item[-] $M(\{P1\})$ has changed, so add the program locations whose fixed point equations directly depend on $M(\{P1\})$ to $W$.
        \begin{itemize}
        \item[-] Add $\{P2\}$ and $\{P7\}$ to $W$.
        \end{itemize}
    \item[-] $W$ is now $\{\{P2\}, \{P3\}, \{P4\}, \{P5\}, \{P6\}, \{P7\}\}$.
    \end{itemize}
\end{itemize}

\begin{itemize}
\item[-] Pick $\{P2\}$ from $W$.
    \begin{itemize}
    \item[-] Remove $\{P2\}$ from $W$.
    \item[-] $M(\{P2\})$ is $\{x : \bot, y : \bot\}$.
    \item[-] Compute $F_2(M)$:
        \begin{itemize}
        \item[-] $M(\{P1\}) \sqcup M(\{P6\}) = \{x : [-\inf, \inf], y : [7, 7]\} \sqcup \{x : \bot, y : \bot\} = \{x : [-\inf, \inf], y : [7, 7]\}$
        \item[-] Filtering $\{x : [-\inf, \inf], y : [7, 7]\}$ by true results in:
            \begin{itemize}
            \item[-] $S = \{x : [-\inf, \inf], y : [7, 7]\}$
            \end{itemize}
        \end{itemize}
    \item[-] Because $\{P2\}$ corresponds to a loop head, we widen $M(\{P2\})$ by $S$.
        \begin{itemize}
        \item[-] $M(\{P2\}) \nabla S$ results in $S' = \{x : [-\inf, \inf], y : [7, 7]\}$, where
            \begin{itemize}
            \item[-] $S'(x) = \bot \nabla [-\inf, \inf]$
                            $= [-\inf, \inf]$
            \item[-] $S'(y) = \bot \nabla [7, 7]$
                            $= [7, 7]$
            \end{itemize}
        \item[-] Update $M(\{P2\})$ to be $\{x : [-\inf, \inf], y : [7, 7]\}$.
        \end{itemize}
    \item[-] $M(\{P2\})$ has changed, so add the program locations whose fixed point equations directly depend on $M(\{P2\})$ to $W$.
        \begin{itemize}
        \item[-] Add $\{P3\}$ to $W$.
        \end{itemize}
    \item[-] $W$ is now $\{\{P3\}, \{P4\}, \{P5\}, \{P6\}, \{P7\}\}$.
    \end{itemize}
\end{itemize}

\begin{itemize}
\item[-] Pick $\{P3\}$ from $W$.
    \begin{itemize}
    \item[-] Remove $\{P3\}$ from $W$.
    \item[-] $M(\{P3\})$ is $\{x : \bot, y : \bot\}$.
    \item[-] Compute $F_3(M)$, resulting in $M(\{P3\}) = \{x : [-\inf, \inf], y : [7, 7]\}$, where
        \begin{itemize}
        \item[-] $M(\{P3\})(x) = [-\inf, \inf]$, which is the result of interpreting $x := \text{read}()$.
        \item[-] $M(\{P3\})(y) = M(\{P2\})(y)$
        \end{itemize}
    \item[-] $M(\{P3\})$ has changed, so add the program locations whose fixed point equations directly depend on $M(\{P3\})$ to $W$.
        \begin{itemize}
        \item[-] Add $\{P4\}$ and $\{P6\}$ to $W$.
        \end{itemize}
    \item[-] $W$ is now $\{\{P4\}, \{P5\}, \{P6\}, \{P7\}\}$.
    \end{itemize}
\end{itemize}

\begin{itemize}
\item[-] Pick $\{P4\}$ from $W$.
   \begin{itemize}
   \item[-] Remove $\{P4\}$ from $W$.
   \item[-] $M(\{P4\})$ is $\{x : \bot, y : \bot\}$
   \item[-] Compute $F_4(M)$:
       \begin{itemize}
       \item[-] $M(\{P3\}) \sqcup M(\{P5\}) = \{x : [-\inf, \inf], y : [7, 7]\} \sqcup \{x : \bot, y : \bot\} = \{x : [-\inf, \inf], y : [7, 7]\}$
       \item[-] Filtering $\{x : [-\inf, \inf], y : [7, 7]\}$ by $x \leq y$ results in:
           \begin{itemize}
           \item[-] $S = \{x : [-\inf, 7], y : [7, 7]\}$.
           \end{itemize}
       \end{itemize}
   \item[-] Because $\{P4\}$ corresponds to a loop head, we widen $M(\{P4\})$ by $S$.
       \begin{itemize}
       \item[-] $M(\{P4\}) \nabla S$ results in $S' = \{x : [-\inf, 7], y : [7, 7]\}$, where
           \begin{itemize}
           \item[-] $S'(x) = \bot \nabla [-\inf, 7]$
                           $= [-\inf, 7]$
           \item[-] $S'(y) = \bot \nabla [7, 7]$
                           $= [7, 7]$
           \end{itemize}
       \item[-] Update $M(\{P4\})$ to be $\{x : [-\inf, 7], y : [7, 7]\}$.
       \end{itemize}
   \item[-] $M(\{P4\})$ has changed, so add the program locations whose fixed point equations directly depend on $M(\{P4\})$ to $W$.
       \begin{itemize}
       \item[-] Add $\{P5\}$ to $W$.
       \end{itemize}
   \item[-] $W$ is now $\{\{P5\}, \{P6\}, \{P7\}\}$.
   \end{itemize}
\end{itemize}

\begin{itemize}
\item[-] Pick $\{P5\}$ from $W$.
   \begin{itemize}
   \item[-] Remove $\{P5\}$ from $W$.
   \item[-] $M(\{P5\})$ is $\{x : \bot, y : \bot\}$.
   \item[-] Compute $F_5(M)$ and update the value of $M(\{P5\})$, resulting in $M(\{P5\}) = \{x : [-\inf, 8], y : [7, 7]\}$, where
       \begin{itemize}
       \item[-] $M(\{P5\})(x) = M(\{P4\})(x) + [1, 1]$
                               $= [-\inf, 7] + [1, 1]$
                               $= [-\inf, 8]$
       \item[-] $M(\{P5\})(y) = M(\{P4\})(y)$
                               $= [7, 7]$
       \end{itemize}
   \item[-] $M(\{P5\})$ has changed, so add the program locations whose fixed point equations directly depend on $M(\{P5\})$ to $W$.
       \begin{itemize}
       \item[-] Add $\{P4\}$ and $\{P6\}$ to $W$.
       \end{itemize}
   \item[-] $W$ is now $\{\{P4\}, \{P6\}, \{P7\}\}$.
   \end{itemize}
\end{itemize}

\begin{itemize}
\item[-] Pick $\{P4\}$ from $W$.
   \begin{itemize}
   \item[-] Remove $\{P4\}$ from $W$.
   \item[-] $M(\{P4\}) = \{x : [-\inf, 7], y : [7, 7]\}$.
   \item[-] Compute $F_4(M)$:
       \begin{itemize}
       \item[-] $M(\{P3\}) \sqcup M(\{P5\}) = \{x : [-\inf, \inf], y : [7, 7]\} \sqcup \{x : [-\inf, 8], y : [7, 7]\} = \{x : [-\inf, \inf], y : [7, 7]\}$
       \item[-] Filtering $\{x : [-\inf, \inf], y : [7, 7]\}$ by $x \leq y$ results in:
           \begin{itemize}
           \item[-] $S = \{x : [-\inf, 7], y : [7, 7]\}$.
           \end{itemize}
       \end{itemize}
   \item[-] Because $\{P4\}$ corresponds to a loop head, we widen $M(\{P4\})$ by $S$.
       \begin{itemize}
       \item[-] $M(\{P4\}) \nabla S$ results in $S' = \{x : [-\inf, 7], y : [7, 7]\}$, where
           \begin{itemize}
           \item[-] $S'(x) = [-\inf, 7] \nabla [-\inf, 7]$
                           $= [-\inf, 7]$
           \item[-] $S'(y) = [7, 7] \nabla [7, 7]$
                           $= [7, 7]$
           \end{itemize}
       \item[-] Update $M(\{P4\})$ to be $\{x : [-\inf, 7], y : [7, 7]\}$.
       \end{itemize}
   \item[-] $M(\{P4\})$ has not changed, so we don't add anything to $W$.
   \item[-] $W$ is now $\{\{P6\}, \{P7\}\}$.
   \end{itemize}
\end{itemize}

\begin{itemize}
\item[-] Pick $\{P6\}$ from $W$.
   \begin{itemize}
   \item[-] Remove $\{P6\}$ from $W$.
   \item[-] $M(\{P6\}) = \{x : \bot, y : \bot\}$.
   \item[-] Compute $F_6(M)$:
       \begin{itemize}
       \item[-] $M(\{P3\}) \sqcup M(\{P5\}) = \{x : [-\inf, \inf], y : [7, 7]\} \sqcup \{x : [-\inf, 8], y : [7, 7]\} = \{x : [-\inf, \inf], y : [7, 7]\}$
       \item[-] Filtering $\{x : [-\inf, \inf], y : [7, 7]\}$ by $x > y$ results in $\{x : [8, \inf], y : [7, 7]\}$.
       \item[-] Update $M(\{P6\})$ to be $\{x : [8, \inf], y : [7, 7]\}$.
       \end{itemize}
   \item[-] $M(\{P6\})$ has changed, so add the program locations whose fixed point equations directly depend on $M(\{P6\})$ to $W$.
       \begin{itemize}
       \item[-] Add $\{P2\}$ and $\{P7\}$ to $W$.
       \end{itemize}
   \item[-] $W$ is now $\{\{P2\}, \{P7\}\}$.
   \end{itemize}
\end{itemize}

\begin{itemize}
\item[-] Pick $\{P2\}$ from $W$.
   \begin{itemize}
   \item[-] Remove $\{P2\}$ from $W$.
   \item[-] $M(\{P2\}) = \{x : [-\inf, \inf], y : [7, 7]\}$.
   \item[-] Compute $F_2(M)$:
       \begin{itemize}
       \item[-] $M(\{P1\}) \sqcup M(\{P6\}) = \{x : [-\inf, \inf], y : [7, 7]\} \sqcup \{x : [8, \inf], y : [7, 7]\} = \{x : [-\inf, \inf], y : [7, 7]\}$
       \item[-] Filtering $\{x : [-\inf, \inf], y : [7, 7]\}$ by true results in:
           \begin{itemize}
           \item[-] $S = \{x : [-\inf, \inf], y : [7, 7]\}$.
           \end{itemize}
       \end{itemize}
   \item[-] Because $\{P2\}$ corresponds to a loop head, we widen $M(\{P2\})$ by $S$.
       \begin{itemize}
       \item[-] $M(\{P2\}) \nabla S$ results in $S' = \{x : [-\inf, \inf], y : [7, 7]\}$, where
           \begin{itemize}
           \item[-] $S'(x) = [-\inf, \inf] \nabla [-\inf, \inf]$
                           $= [-\inf, \inf]$
           \item[-] $S'(y) = [7, 7] \nabla [7, 7]$
                           $= [7, 7]$
           \end{itemize}
       \item[-] Update $M(\{P2\})$ to be $\{x : [-\inf, \inf], y : [7, 7]\}$.
       \end{itemize}
   \item[-] $M(\{P2\})$ has not changed, so don't add anything to $W$.
   \item[-] $W$ is now $\{\{P7\}\}$.
   \end{itemize}
\end{itemize}

\begin{itemize}
\item[-] Pick $\{P7\}$ from $W$.
   \begin{itemize}
   \item[-] Remove $\{P7\}$ from $W$.
   \item[-] $M(\{P7\}) = \{x : \bot, y : \bot\}$.
   \item[-] Compute $F_7(M)$
       \begin{itemize}
       \item[-] $M(\{P1\}) \sqcup M(\{P6\}) = \{x : [-\inf, \inf], y : [7, 7]\} \sqcup \{x : [8, \inf], y : [7, 7]\} = \{x : [-\inf, \inf], y : [7, 7]\}$.
       \item[-] Filtering $\{x : [-\inf, \inf], y : [7, 7]\}$ by false results in:
           \begin{itemize}
           \item[-] $S = \{x : \bot, y : \bot\}$.
           \end{itemize}
       \item[-] Update $M(\{P7\})$ to be $\{x : \bot, y : \bot\}$.
       \end{itemize}
   \item[-] $M(\{P7\})$ has not changed, so don't add anything to $W$.
   \item[-] $W$ is now $\{\}$.
   \end{itemize}
\end{itemize}

The worklist is empty, meaning we've finished the analysis and M is

\[
M(\{P_0\}) = \{x : [-\inf, \inf], y : [-\inf, \inf]\}
\]
\[
M(\{P_1\}) = \{x : [-\inf, \inf], y : [7, 7]\}
\]
\[
M(\{P_2\}) = \{x : [-\inf, \inf], y : [7, 7]\}
\]
\[
M(\{P_3\}) = \{x : [-\inf, \inf], y : [7, 7]\}
\]
\[
M(\{P_4\}) = \{x : [-\inf, 7], y : [7, 7]\}
\]
\[
M(\{P_5\}) = \{x : [-\inf, 8], y : [7, 7]\}
\]
\[
M(\{P_6\}) = \{x : [8, \inf], y : [7, 7]\}
\]
\[
M(\{P_7\}) = \{x : \bot, y : \bot\}
\]

Now, please solve this, outputting the intermediary steps you take:

\textbf{[Input Program]}
\end{tcolorbox}
\end{document}